\documentclass[sigconf]{acmart}
\usepackage{multirow}

\usepackage{amsmath}
\usepackage{algorithm} 
\usepackage{algpseudocode}

\usepackage{array}
\newcolumntype{x}[1]{>{\centering\arraybackslash\hspace{0pt}}p{#1}}

\usepackage{amsmath,amsfonts,bm}

\def\eqref#1{equation~\ref{#1}}

\def\1{\bm{1}}

\def\rr{{\textnormal{r}}}

\def\mP{{\bm{P}}}

\def\mR{{\bm{R}}}

\def\mT{{\bm{T}}}

\def\mX{{\bm{X}}}

\DeclareMathAlphabet{\mathsfit}{\encodingdefault}{\sfdefault}{m}{sl}
\SetMathAlphabet{\mathsfit}{bold}{\encodingdefault}{\sfdefault}{bx}{n}

\newcommand{\R}{\mathbb{R}}

\usepackage{hyperref}
\usepackage{url}

\usepackage{booktabs}
\usepackage{graphicx}
\usepackage{multirow}
\usepackage{algorithm} 
\usepackage{algpseudocode} 
\usepackage[english]{babel}
\usepackage{amsthm}
\newtheorem{theorem}{Theorem}

\newtheorem{lemma}{Lemma}

\copyrightyear{2024}
\acmYear{2024}
\setcopyright{acmlicensed}\acmConference[KDD '24]{Proceedings of the 30th ACM SIGKDD Conference on Knowledge Discovery and Data Mining}{August 25--29, 2024}{Barcelona, Spain}
\acmBooktitle{Proceedings of the 30th ACM SIGKDD Conference on Knowledge Discovery and Data Mining (KDD '24), August 25--29, 2024, Barcelona, Spain}
\acmDOI{10.1145/3637528.3671808}
\acmISBN{979-8-4007-0490-1/24/08}



\begin{document}

\title{Understanding Inter-Session Intentions via Complex Logical Reasoning}

\author{Jiaxin Bai}
\authornote{~~~Work done during an internship at Amazon.}
\affiliation{%
\institution{Department of CSE, HKUST}
 \country{Hong Kong SAR, China}
}
\email{jbai@connect.ust.hk}

\author{Chen Luo}
\affiliation{%
  \institution{Amazon.com Inc}
  \city{Palo Alto}
  \country{USA}}
\email{cheluo@amazon.com}

\author{Zheng Li}
\affiliation{%
  \institution{Amazon.com Inc}
  \city{Palo Alto}
  \country{USA}}
\email{amzzhe@amazon.com}

\author{Qingyu Yin}
\affiliation{%
  \institution{Amazon.com Inc}
  \city{Palo Alto}
  \country{USA}}
\email{qingyy@amazon.com}

\author{Yangqiu Song}
\authornote{~~~Visiting academic scholar at Amazon.}
\affiliation{%
  \institution{Department of CSE, HKUST}
 \country{Hong Kong SAR, China}
}
\email{yqsong@cse.ust.hk}

\begin{abstract}

Understanding user intentions is essential for improving product recommendations, navigation suggestions, and query reformulations. However, user intentions can be intricate, involving multiple sessions and attribute requirements connected by logical operators such as And, Or, and Not. For instance, a user may search for Nike or Adidas running shoes across various sessions, with a preference for purple. In another example, a user may have purchased a mattress in a previous session and is now looking for a matching bed frame without intending to buy another mattress. Existing research on session understanding has not adequately addressed making product or attribute recommendations for such complex intentions.
In this paper, we present the task of logical session complex query answering (LS-CQA), where sessions are treated as hyperedges of items, and we frame the problem of complex intention understanding as an LS-CQA task on an aggregated hypergraph of sessions, items, and attributes. This is a unique complex query answering task with sessions as ordered hyperedges. We also introduce a new model, the Logical Session Graph Transformer (LSGT), which captures interactions among items across different sessions and their logical connections using a transformer structure.
We analyze the expressiveness of LSGT and prove the permutation invariance of the inputs for the logical operators. By evaluating LSGT on three datasets, we demonstrate that it achieves state-of-the-art results.

\end{abstract}

\maketitle
\section{Introduction}

Understanding user intention is a critical challenge in product search. 
A user's intention can be captured in many ways during the product search process. 
Some intentions can be explicitly given through search keywords. 
For example, a user may use keywords like ``Red Nike Shoes'' to indicate the desired product type, brand, and color. 
However, search keywords may not always accurately reflect the user's intention, especially when they are unsure of what they want initially. 
To address this issue, session-based recommendation methods have been proposed to leverage user behavior information to make more accurate recommendations \citep{hidasi2015session, li2017neural}.

\begin{figure*}[t]
\begin{center}
\includegraphics[width=0.8\linewidth]{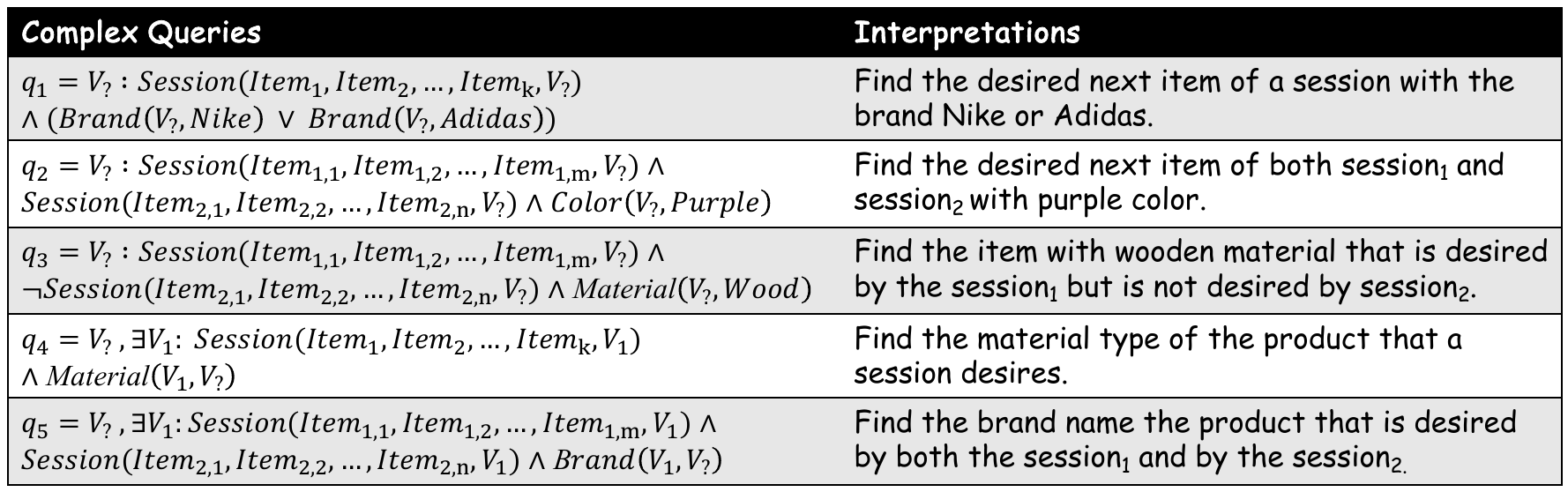}
\end{center}
\vspace{-0.4cm}
\caption{
Example complex queries involving varied numbers of sessions, products, and product attributes.
}
\label{fig:query_examples}
\end{figure*}

User intentions are usually complex. Users often have several explicit requirements for desired items, such as brand names, colors, sizes, and materials. 
For example, in Figure~\ref{fig:query_examples}, query $q_1$ shows a user desiring Nike or Adidas products in the current session. On the other hand, users may spend multiple sessions before making a purchasing decision. 
For query $q_2$, the user spends two sessions searching for a desired product with an explicit requirement of purple color. 
Moreover, these requirements can involve logical structures. 
For instance, a user explicitly states that they do not want products similar to a previous session. 
In query $q_3$, a user has purchased a mattress in a previous session and is now looking for a wooden bed frame, without any intention of buying another mattress. 
With the help of logical operators like \texttt{AND} $\land$, \texttt{OR} $\lor$, and \texttt{NOT} $\lnot$, we can describe the complex intentions by using a complex logical session query, like $q_1$, $q_2$, and $q_3$ in Figure~\ref{fig:query_examples}.

Furthermore, there are scenarios where we are interested in obtaining product attributes based on sessions. For example, in query $q_4$ shown in Figure \ref{fig:query_examples}, we aim to identify the material types of the products desired in the session. Similarly, in query $q_5$, when given two sessions, we want to determine the brand names of the products desired in both sessions.
To address these scenarios, we can describe these queries using logic expressions and variables. For instance, we can use the variable $V_1$ to represent the products and $V_?$ to represent the attribute associated with the product $V_1$.
As a result, the task of recommending attributes based on complex user intentions can be formulated as logical query answering. We inquire about the attribute $V_?$ such that there exists a product $V_1$ in the given sessions, and the product attribute $V_?$ is desired.

To systematically answer queries with complex user intentions, we formally propose the task of logical session complex query answering (LS-CQA). This can be seen as an extension of the complex query answering (CQA) problem to relational hypergraph data, where sessions are treated as ordered hyperedges of items. The task of product or attribute recommendation under complex intention is reformulated as a task of answering logical queries on an aggregated hypergraph of sessions, items, and attributes. Figure \ref{fig:hypergraph_types} (C) provides an example of such an example, where each session is represented as a hyperedge connecting the corresponding items.

In addition to utilizing CQA methods with N-ary facts, such as NQE proposed by \cite{DBLP:conf/aaai/LuoEYZGYTLW23}, another more reasonable approach to  LS-CQA is to employ a session encoder. Recent studies \citep{DBLP:conf/sigir/LiPLSLXYCZZ21, DBLP:conf/icml/HuangWLH22, DBLP:conf/wsdm/ZhangGLXKZXWK23} 
have shown the effectiveness of session encoders in encoding sessions and generating session representations.
However, the neural session encoders tend to conduct implicit abstraction of products during the session encoding process \citep{DBLP:conf/wsdm/ZhangGLXKZXWK23}. 
The logical query encoder can only access the abstracted session representations, resulting in a lack of capturing the interactions between items in different sessions during the query encoding.

Motivated by this, we introduce the Logical Session Graph Transformer (LSGT) as an approach for encoding complex query sessions as hypergraphs\footnote{Code available: https://github.com/HKUST-KnowComp/SessionCQA}. 
Building upon the work by \cite{DBLP:conf/nips/KimNMCLLH22}, we transform items, sessions, relation features, session structures, and logical structures into tokens, and they are then encoded using a standard transformer model. This transformation enables us to effectively capture interactions among items in different sessions through the any-to-any attention mechanisms in transformer models.
By analyzing the Relational Weisfeiler-Lehman by \cite{DBLP:conf/log/Barcelo00O22, DBLP:journals/corr/abs-2302-02209}, we provide theoretical justification for LSGT, demonstrating that it possesses the expressiveness of at least 1-RWL, and has at least same expressiveness as existing logical query encoders that employ message-passing mechanisms for logical query encoding in WL test. 
Meanwhile, LSGT maintains the property of operation-wise permutation invariance, similar to other logical query encoders.
To evaluate LSGT, we have conducted experiments on three evaluation datasets: Amazon, Diginetica, and Dressipi. 
Results demonstrate that LSGT achieves state-of-the-art performance on these datasets.
In general, the contribution of this paper can be summarized as follows: 

\begin{itemize}
    \item We extend complex query answering (CQA) to hypergraphs with sessions as ordered hyperedges (LS-CQA) for describing and solving the product and attribute recommendations with complex user intentions. We also constructed three corresponding scaled datasets with the full support of first-order logical operators (intersection, union, negation) for evaluating CQA models on hypergraphs with ordered hyperedges and varied arity. 

    \item We propose a new method, logical session graph transformer (LSGT). We use tokens and identifiers to uniformly represent the items, sessions, logical operators, and their relations. Then we use a transformer structure to encode them. 

    \item  We conducted experiments on Amazon, Digintica, and Dressipi to show that existing Transformer-based models show similar results on 3 benchmarks despite different linearization strategies. Meanwhile, We also find the linearization of LSGT leads to improvements in queries with negations and unseen query types. Meanwhile,  We theoretically justify the expressiveness in the Weisfeiler-Lehman (WL) test and the Relational Weisfeiler-Lehman (RWL) test. We also prove the operator-wise permutation invariance of LSGT. 
\end{itemize}

\section{Problem Formulation}

\subsection{logical session complex query Answering }

In previous work, complex query answering is usually conducted on a knowledge graph $\mathcal{G} = (\mathcal{V}, \mathcal{R})$. 
However, on our aggregated hypergraph, there are items, sessions, and attribute values. Because of this, the graph definition is $\mathcal{G} = (\mathcal{V}, \mathcal{R}, \mathcal{S})$. 
The $\mathcal{V}$ is the set of vertices $v$, and the $\mathcal{R}$ is the set of relation $r$. 
The $\mathcal{S}$ is the set of sessions regarded as directed hyperedges. 
To describe the relations in logical expressions, the relations are defined in functional forms. 
Each relation $r$ is defined as a function, and each relation has two arguments, which are two items or attributes $v$ and $v'$. 
The value of function $r(v, v') = 1$ if and only if there is a relation between the items or attributes $v$ and $v'$.
Each session $s \in \mathcal{S}$ is the sequence of vertices where $s(v_1, v_2,..., v_n) = 1$ if and only if $v_1, v_2,..., v_n$ appeared in the same session. 

\begin{figure*}[t]
\begin{center}
\includegraphics[width=0.9\linewidth]{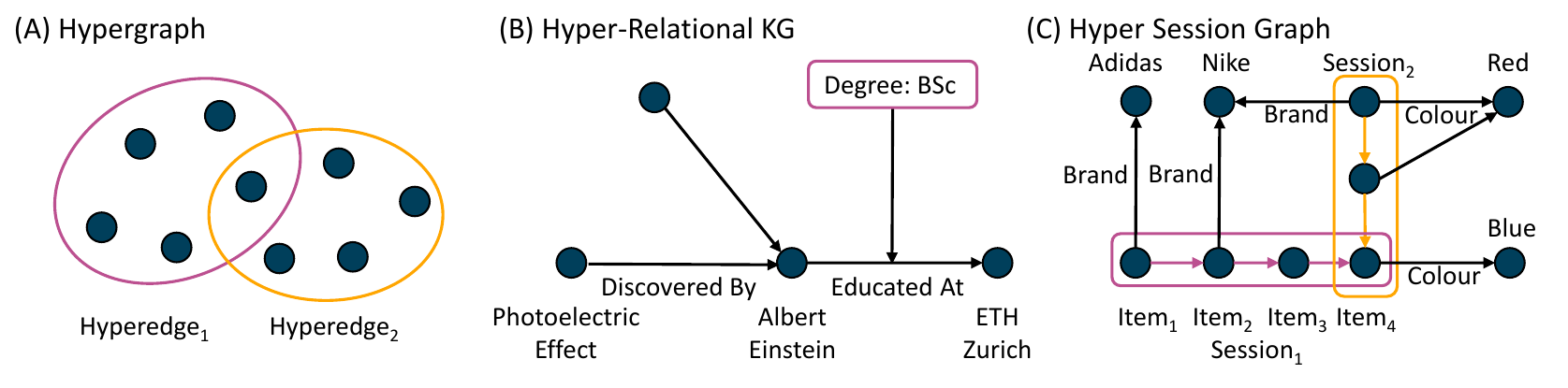}
\end{center}
\vspace{-0.3cm}
\caption{
This figure shows the connections and differences between general hypergraphs, hyper-relational knowledge graphs, and the hyper-session graph in our problem. 
}
\label{fig:hypergraph_types}
\end{figure*}

\begin{figure*}[t]
\begin{center}
\includegraphics[width=0.9\linewidth]{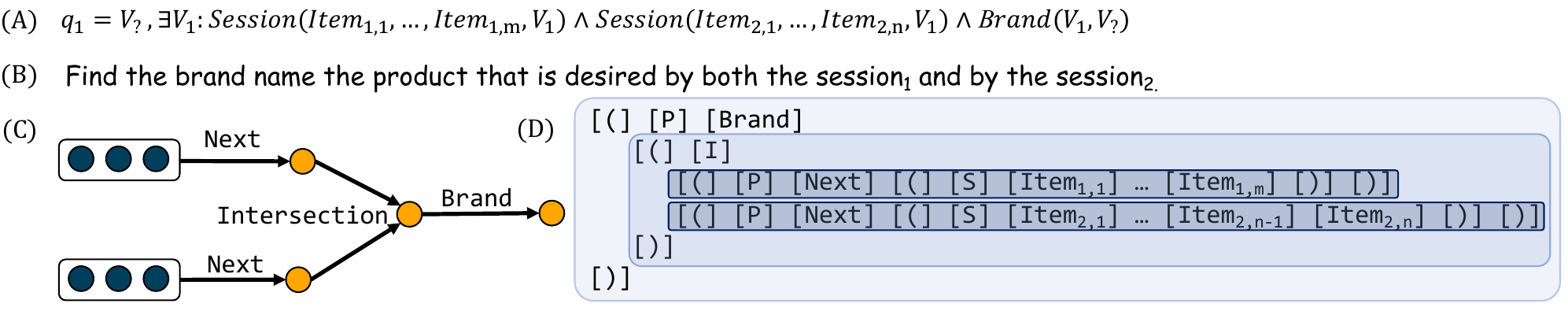}
\end{center}
\vspace{-0.4cm}
\caption{
This figure shows the illustration of different query embedding methods. (A) The logical session complex query is expressed in the first-order logic form. (B) The interpretations on the logical session complex query. (C) The computational graph of the complex query proposed by \cite{hamilton2018embedding}; (D) The linearization of the computational graph to token proposed by \cite{bai-etal-2022-query2particles}.}  
\label{fig:query_illustration}
\vspace{-0.3cm}
\end{figure*}

The queries are defined in the first-order logical (FOL) forms. 
In a first-order logical expression, there are logical operations such as existential quantifiers $\exists$, conjunctions $\land$, disjunctions $\lor$, and negations $\lnot$. 
In such a logical query, there are anchor items or attribute $V_a \in \mathcal{V}$, existential quantified variables $V_1, V_2, ... V_k \in \mathcal{V}$, and a target variable $V_?\in \mathcal{V}$. 
The knowledge graph query is written to find the answer $V_?\in \mathcal{V}$, such that there exist $V_1, V_2, ... V_k \in \mathcal{V}$ satisfying the logical expression in the query. 
For each query, it can be converted to a disjunctive normal form, where the query is expressed as a disjunction of several conjunctive expressions:
\begin{align}
q[V_?] &= V_? . \exists V_1, ..., V_k: c_1 \lor c_2 \lor ... \lor c_n ,\label{equa:definition} \\ 
c_i &= e_{i1} \land e_{i2} \land ... \land e_{im} .
\end{align}
Each $c_i $ represents a conjunctive expression of  literals $e_{ij}$, and each
$e_{ij}$ is an atomic or the negation of an atomic expression in any of the following forms:
$e_{ij} = r(v_a, V)$, 
$e_{ij} = \lnot r(v_a, V)$, 
$e_{ij} = r(V, V')$, or 
$e_{ij} = \lnot r(V, V')$.
The atomics $e_{ij}$ can be also hyper N-ary relations between vertices indicating that there exists a session among them. 
In this case, the $e_{ij} = s(v_1, v_2, ..., v_n, V)$ or its negations $e_{ij} = \lnot s(v_1, v_2, ..., v_n, V)$.
Here $v_a$ and $v_i \in V_a$ is one of the anchor nodes, and $V,V' \in \{ V_1, V_2, ... ,V_k, V_? \}$ are distinct variables satisfying $V \neq V'$. 
When a query is an existential positive first-order (EPFO) query, there are only conjunctions $\land$ and disjunctions $\lor$ in the expression (no negations $\lnot$). When the query is a conjunctive query, there are only conjunctions $\land$ in the expressions (no disjunctions $\lor$ and negations $\lnot$).

\section{Related Work}

\subsection{Hyper-Relational Graph Reasoning}

The reasoning over hyper-relational KG proposed by  \cite{DBLP:conf/iclr/AlivanistosBC022}, they extend the multi-hop reasoning problem to hyper-relational KGs and propose a method, StarQE, to embed and answer hyper-relational conjunctive queries using Graph Neural Networks and query embedding techniques.
The StarQE conducts message-passing over the quantifier of the hyper-relations in KG, which cannot be directly used for encoding the hyper-relations in this task.
\cite{DBLP:conf/aaai/LuoEYZGYTLW23} propose a novel Nary Query Embedding (NQE) model for complex query answering over hyper-relational knowledge graphs, which can handle more general n-ary FOL queries including existential quantifiers, conjunction, disjunction, and negation.
The encoder design of NQE is more general to N-ary facts, thus it can be directly used for encoding the sessions as hyper-edges.

\begin{figure*}[t]
\begin{center}
\includegraphics[width=\linewidth]{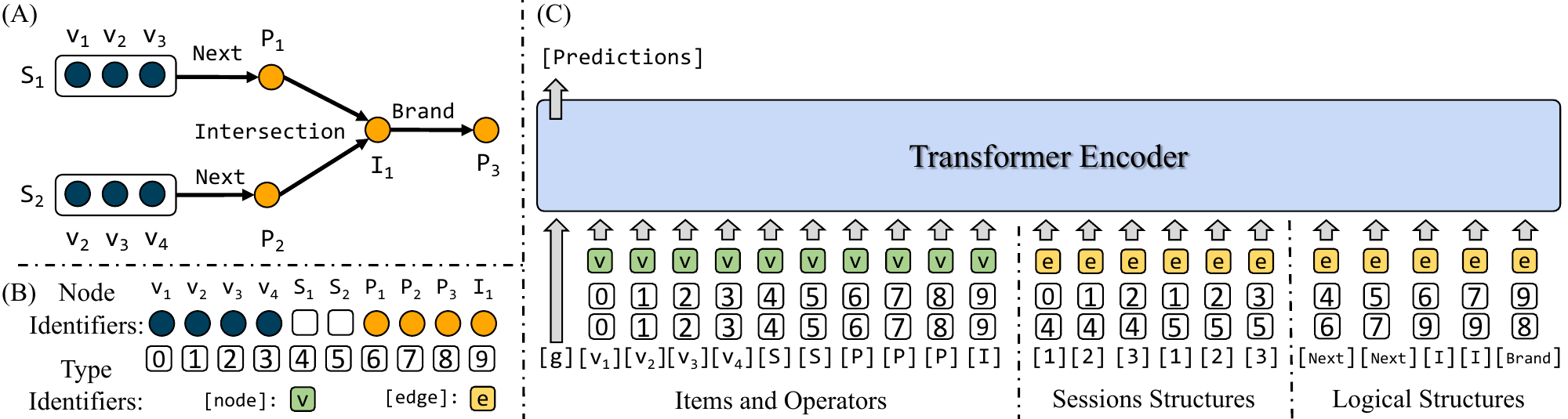}
\end{center}
\vspace{-0.25cm}
\caption{
This figure shows the method of LSGT. (A) The computational graph indicates finding the brand of the products that are designed by both session $S_1$ and $S_2$. (B) The node identifiers and type identifiers for the tokens and each of the identifiers is associated with its corresponding embedding vector. (C) The transformer encoder is used for encoding the tokens.
}
\label{fig:sessiongt}
\end{figure*}
\vspace{-0.1cm}
\subsection{Complex Query Answering}

Previous literature on logical query answering mainly focuses on knowledge graphs~\citep{hamilton2018embedding, ren2020query2box, arakelyan2020complex,DBLP:conf/kdd/BaiLLYYS23,DBLP:conf/nips/BaiLW0S23,DBLP:journals/corr/abs-2312-13866,DBLP:journals/corr/abs-2312-15591,DBLP:journals/corr/abs-2307-13701,DBLP:conf/acl/WangFYSWS23,DBLP:journals/corr/abs-2402-14609,DBLP:conf/www/LiuWBST24}. 
Various methods are proposed to deal with the incompleteness issue of knowledge graphs by using the existing facts in the KG to generalize to the facts that are not in the KG but highly likely to be true. 
Recent advances in query embedding \citep{hamilton2018embedding, ren2020query2box, ren2020beta} methods have shown promise in answering logical queries on large-scaled graph-structured data effectively and efficiently. 
However, they cannot be directly used for answering queries with hyperedges, or in other words N-ary facts.
Meanwhile, there are also methods \citep{DBLP:conf/iclr/AlivanistosBC022, DBLP:conf/aaai/LuoEYZGYTLW23} that can perform robust reasoning on hyper-relational knowledge graphs, which is illustrated in Figure \ref{fig:hypergraph_types} (B). 
Because of the fundamental differences between the hyper-relational knowledge graphs and the hypergraphs of sessions, not all of them can be directly adopted for this task.
Recently, there is also new progress on query encoding that is orthogonal to this paper, which puts a focus on the neural encoders for complex queries.    
\citet{xu2022neural} propose a neural-symbolic entangled method, ENeSy, for query encoding. 
\citet{yang2022gammae} propose to use Gamma Embeddings to encode complex logical queries. 
\citet{liu2022mask} propose to use pre-training on the knowledge graph with kg-transformer and then conduct fine-tuning on the complex query answering.
Meanwhile, query decomposition \citep{arakelyan2020complex} is another way to deal with the problem of complex query answering.
In this method, the probabilities of atomic queries are first computed by a link predictor, and then continuous optimization or beam search is used to conduct inference time optimization. 
Moreover, \cite{DBLP:journals/corr/abs-2301-08859} propose an alternative to query encoding and query decomposition, in which they conduct message passing on the one-hop atomics to conduct complex query answering.
Recently a novel neural search-based method QTO \citep{bai2022answering} is proposed. QTO demonstrates impressive performance CQA. 
There are also neural-symbolic query encoding methods proposed \citep{xu2022neural, zhu2022neural}. In this line of research, their query encoders refer back to the training knowledge graph to obtain symbolic information from the graph. 
LogicRec \citep{DBLP:conf/sigir/TangFTP0S23} discusses recommending highly personalized items based on complex logical requirements, which current recommendation systems struggle to handle.

\subsection{Session Encoders}

In recent literature, various methods have been proposed to reflect user intentions and build better recommendation systems using session history. 
Because of the nature of sequence modeling, various methods utilize recurrent neural networks (RNNs) and convolutions neural networks (CNNs) to model session data \citep{hidasi2015session,li2017neural, DBLP:conf/kdd/LiuZMZ18, DBLP:conf/wsdm/TangW18}. 
Recent developments in session-based recommendation have focused on using Graph Neural Networks (GNNs) to extract relationships and better model transitions within sessions \citep{DBLP:conf/sigir/LiPLSLXYCZZ21,DBLP:conf/cikm/GuoZL0ZK22,DBLP:conf/icml/HuangWLH22}. \citet{DBLP:conf/aaai/WuT0WXT19} were the first to propose using GNNs to capture complex transitions with graph structures, and subsequent research has incorporated position and target information, global context, and highway networks to further improve performance~\citep{DBLP:conf/cikm/PanCCCR20,DBLP:conf/aaai/0013YYWC021}. 
However, previous efforts have focused more on the message-passing part and less on designing effective readout operations to aggregate embeddings to the session-level embedding. 
According to \cite{DBLP:conf/wsdm/ZhangGLXKZXWK23}, current readout operations have limited capacity in reasoning over sessions, and the performance improvement of GNN models is not significant enough to justify the time and memory consumption of sophisticated models. So \cite{DBLP:conf/wsdm/ZhangGLXKZXWK23} proposed a pure attention-based method Atten-Mixer to conduct session recommendations.

\section{Logical Session Graph Transformer}

In this session, we describe the logical session graph transformer (LSGT) for encoding logical queries involving sessions.
In LSGT, the node and edge features, session structures, and logical structures are all converted into tokens and identifiers. Subsequently, they serve as input to a standard transformer encoder model.

\subsection{Items, Sessions, and Operators Tokens}

The first step in LSGT involves assigning node identifiers to items, sessions, and operators. For instance, in Figure \ref{fig:sessiongt}, there are two sessions, $S_1$ and $S_2$, with items $[v_1, v_2, v_3]$ and $[v_2, v_3, v_4]$, respectively. The computational graph then uses relational projection operators $P_1$ and $P_2$ to find the two sets of next items desired by $S_1$ and $S_2$, respectively. Once all items, sessions, and operators have been identified, each is assigned a unique node identifier. For example, $v_1$ to $v_4$ are assigned identifiers from $0$ to $3$, $S_1$ and $S_2$ are assigned identifiers $4$ and $5$, projections from $P_1$ to $P_3$ are assigned identifiers from $6$ to $8$, and intersection operation $I_1$ is assigned to $9$.

In general, when there are $n$ nodes denoted by their identifiers as $\{p_0, p_1, ..., p_n\}$, their node features are assigned as follows: 
if $p_i$ is an item, its features are assigned to its item embedding.
If $p_i$ is a session $S_j$, it is assigned an embedding of \texttt{[S]}. 
If $p_i$ is a neural operator, it is assigned the operator embedding from \texttt{[P]}, \texttt{[I]}, \texttt{[N]}, or \texttt{[U]} based on its operation type. 
The feature matrix for these $n$ nodes is then denoted as $\mX_p \in \R^{n\times d_1}$. 
Additionally, each node identifier is associated with random orthonormal vectors \citep{DBLP:conf/nips/KimNMCLLH22}, denoted as $\mP_p \in \R^{n\times d_2}$. 
All nodes are assigned the type identifier of \texttt{[node]}, which means that they are the nodes in the computational graph. The token type embedding for vertices is denoted as $T_{[node]} \in \R^{d_3}$. 
The input vectors for the transformer are concatenations of node features, the random orthonormal vectors, and token type embeddings, where node identifiers vectors are repeated twice: $ \mX_{u}^{v} = [\mX_p, \mP_p, \mP_p, \mT_{[node]}] \in \R^{n \times (d_1+ 2d_2 + d_3)}$.

\subsection{Session Structure Tokens}

In this part, we describe the process of constructing the input tokens to indicate the session structure, namely which items are in which session in which position. 
Suppose the item $p$ is from session $q$ and at the position of $r$, and there are $m$ item-session correspondences in total. 
First, we use positional encoding $Pos(\rr) \in \R ^{d_1}$ to describe the positional information. 
Meanwhile, as the item and sessions are associated with their node identifiers $p$ and $q$, we use the node identifier vectors $P_p \in \R^{d_2}$ and $P_q \in \R^{d_2}$  to represent them. 
Meanwhile, this token represents a correspondence between two nodes, so we use the \texttt{[edge]} token type embedding to describe this $T_{[edge]} \in \R^{d_3}$. 
As there are in total $m$ of item-session correspondences, we concatenate them together to obtain the input vectors for the tokens representing session structures:  $\mX_{(p,q,r)}^{s} = [Pos(\rr), \mP_p, \mP_q, \mT^{[edge]}] \in \R^{m \times (d_1 + 2d_2 + d_3)} $. 

\subsection{Logical Structure Tokens}
In this part, we describe the process of constructing the input for tokens to indicate the logical structures. 
As shown in Figure \ref{fig:sessiongt}, in an edge representing a logical operation, there are two nodes $p$ and $q$ respectively. 
If the logical operation is projection, then the edge feature is assigned with relation embedding \texttt{[Rel]}. 
Otherwise, the edge feature is assigned with the operation embedding from \texttt{[P]}, \texttt{[I]}, \texttt{[N]}, and \texttt{[U]} accordingly. 
The edge feature is denoted as $R_r \in \R ^{d_1}$.
Similarly, we use the node identifier vectors $P_p \in \R^{d_2}$ and $P_q \in \R^{d_2}$  to represent involved nodes $p$ and $q$.
Meanwhile, this token represents an edge in the computational graph, so we also associate it with token type embedding $T_{[edge]} \in \R^{d_3}$ to describe it.
Suppose there are in total $w$ such logical edges, we concatenate them together to obtain the input vectors for the tokens representing logical structure: $\mX_{(p,q,r)}^{l} = [\mR_r, \mP_p, \mP_q, \mT_{[edge]}] \in \R^{w \times (d_1 + 2d_2 + d_3)} $.

\subsection{Training LSGT}

After obtaining the three parts describing the items, session structures, and logical structures, we concatenate them together $\mX = [X_{[\texttt{graph}]}, \mX^{v}, \mX^{s}, \mX^{l}] \in R^{(m+n+w+1) \times (d_1+2d_2+d_3)}$, and use this matrix as the input for a standard transformer encoder for compute the query encoding of this complex logical session query. 
Then we append a special token \texttt{[graph]} with embedding $X_{[\texttt{graph}]} \in \R^{d_1+2d_2+d_3} $ at the beginning of the transformer and use the token output of the \texttt{[graph]} token as the embedding of the complex logical session query. 
To train the LSGT model, we compute the normalized probability of the vertice $a$ being the correct answer of query $q$ by using the \texttt{softmax} function on all similarity scores,
\begin{align}
    p(q, a) = \frac{e^{<e_q, e_a>}}{\sum_{a'\in V} e^{<e_q, e_{a'}>}  }.
\end{align}
Then we construct a cross-entropy loss  to maximize the log probabilities of all correct pairs:
\begin{align}
    L = -\frac{1}{N} \sum_i  \log p(q^{(i)}, a^{(i)}).
\end{align}
Each $(q^{(i)}, a^{(i)})$ denotes one of the positive query-answer pairs, and there are $N$ pairs.

\begin{table}[t]
\centering
\caption{The statistics of the constructed hypergraph on sessions, items, and their attribute values are shown.} \label{tab:kg_details} 

\small
\begin{tabular}{@{}p{0.9cm}p{0.8cm}p{0.8cm}p{0.8cm}p{0.8cm}p{0.9cm}p{0.8cm}@{}}
\toprule
Dataset &  Edges& Vertices&  Sessions &  Items &  Attributes & Relations \\ \midrule
Amazon & 8,004,984 & 2,431,747 & 720,816 & 431,036 & 1,279,895 & 10 \\
Diginetica & 1,387,861 & 266,897 & 12,047 & 134,904 & 125,204 & 3 \\
Dressipi & 2,698,692 & 674,853 & 668,650 & 23,618 & 903 & 74 \\ \bottomrule
\end{tabular}
\end{table}

\subsection{Theoretical Properties of LSGT}

In this part, we analyze the theoretical properties of LSGT, focusing on two perspectives.  First, we analyze the expressiveness of LSGT compared to baseline methods in Theorem \ref{theorem:expressiveness} and \ref{theorem:expressiveness_RWL}. Second, we analyze whether LSGT has operator-wise permutation invariant, and this is important in query encoding as operators like \texttt{Intersection} and \texttt{Union} are permutation invariant to inputs in Theorem \ref{theorem:permutation_invariance}. 
We prove the following theorems in the Appendix \ref{sess:proof} of LSGT: 
\begin{theorem}
\label{theorem:expressiveness}
When without considering the relation types in the query graph, the expressiveness of the LSGT encoder is at least the same as that of the encoder that combines a session encoder followed by a logical query encoder under Weisfeiler-Lehman tests\citep{DBLP:conf/iclr/MaronBSL19}.
\end{theorem}

\begin{theorem}
\label{theorem:expressiveness_RWL}
When considering the query graphs are multi-relational graphs with edge relation types, the expressiveness of the LSGT encoder is also at least as powerful as 1-RWL, namely the expressiveness of R-GCN and CompGCN~\citep{DBLP:conf/log/Barcelo00O22,DBLP:journals/corr/abs-2302-02209}.
\end{theorem}

\begin{theorem}
\label{theorem:permutation_invariance}
LSGT can approximate a logical query encoding model that is operator-wise input permutation invariant. 
\end{theorem}

\begin{figure}[t]
\begin{center}
\includegraphics[width=\linewidth]{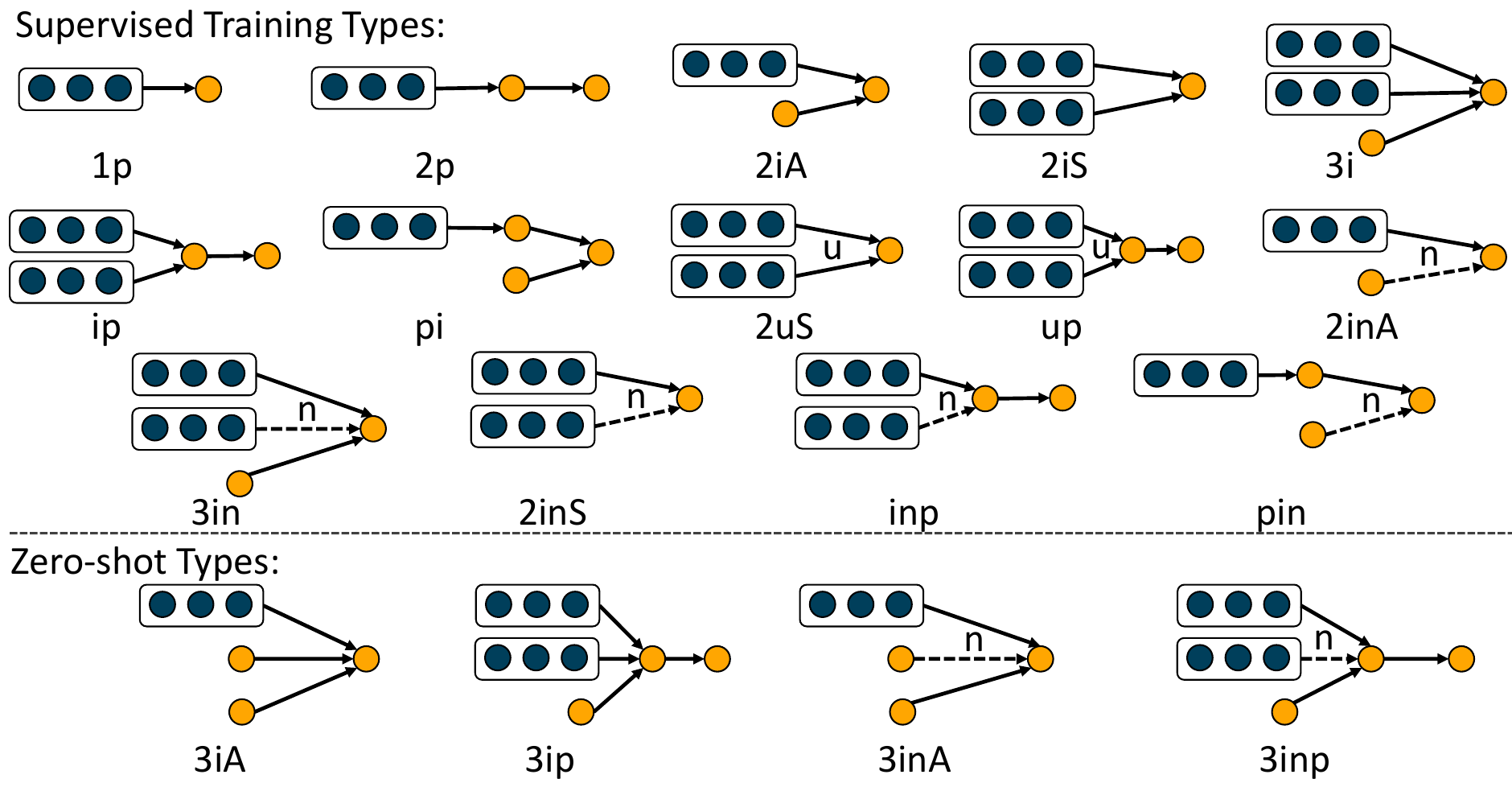}
\end{center}
\caption{
The query structures are used for training and evaluation. For brevity, the $p$, $i$, $n$, and $u$ represent the projection, intersection, negation, and union operations. The query types are trained and evaluated under supervised settings. 
}
\label{fig:query_types}
\end{figure}

\begin{table*}[t]
\centering
\small
\caption{The performance in the mean reciprocal ranking of LSGT compared with the baseline models of SQE and NQE with different backbone modules. Statistical significance is denoted by *.}
\label{tab:main_result}
\vspace{-0.1cm}
\begin{tabular}{@{}l|l|l|p{0.6cm}p{0.6cm}p{0.6cm}p{0.6cm}p{0.6cm}p{0.6cm}p{0.6cm}p{0.6cm}p{0.6cm}|l@{}}
\toprule
Dataset & Query Encoder & Session Encoder & 1p & 2p & 2ia & 2is & 3i & pi & ip & 2u & up & Average EPFO \\ \midrule
\multirow{10}{*}{Amazon} & \multirow{3}{*}{FuzzQE} & GRURec & 11.99 & 6.96 & 54.79 & 88.47 & 68.13 & 16.73 & 14.49 & 10.02 & 6.87 & 30.94 \\
 &  & SRGNN & 13.52 & 7.93 & 54.12 & 85.45 & 67.56 & 18.62 & \textbf{20.46} & 10.82 & 7.24 & 31.75 \\
 &  & Attn-Mixer & 16.79 & 7.96 & 55.76 & \textbf{89.64} & 69.16 & 14.87 & 9.93 & 13.83 & 7.20 & 31.68 \\ \cmidrule(l){2-13} 
 & \multirow{3}{*}{Q2P} & GRURec & 11.60 & 6.83 & 34.73 & 67.64 & 42.18 & 16.66 & 13.82 & 8.54 & 5.82 & 23.09 \\
 &  & SRGNN & 13.94 & 7.69 & 35.89 & 69.90 & 44.61 & 16.19 & 16.44 & 10.20 & 6.46 & 24.59 \\
 &  & Attn-Mixer & 15.93 & 8.53 & 46.67 & 68.62 & 61.13 & 16.95 & 15.78 & 12.43 & 7.41 & 28.16 \\ \cmidrule(l){2-13} 
 & NQE & - & 5.60 & 2.50 & 48.40 & 77.98 & 63.06 & 2.20 & 1.80 & 4.20 & 3.00 & 23.19 \\
 & SQE-Transformer & - & 16.09 & 8.30 & 53.90 & 72.26 & 64.48 & 17.54 & 16.80 & 13.86 & 7.35 & 30.07 \\
 & SQE-LSTM & - & 16.59 & 7.45 & 55.60 & 86.81 & 69.11 & 17.86 & 19.04 & 13.46 & 6.87 & 32.53 \\ \cmidrule(l){2-13} 
 & LSGT (Ours) &  & \textbf{17.73*} & \textbf{9.10*} & \textbf{56.73*} & 84.62 & \textbf{69.39} & \textbf{19.39*} & 19.47 & \textbf{15.40*} & \textbf{7.86*} & \textbf{33.26* (+0.73)} \\ \midrule
\multirow{10}{*}{Diginetica} & \multirow{3}{*}{FuzzQE} & GRURec & 24.10 & 12.29 & 82.48 & 89.19 & 86.26 & 11.64 & 23.34 & 18.19 & 11.18 & 39.85 \\
 &  & SRGNN & 22.53 & 12.33 & 83.19 & 88.35 & 86.26 & 12.55 & 29.56 & 19.76 & 11.48 & 40.67 \\
 &  & Attn-Mixer & \textbf{33.87} & 11.89 & 82.94 & 88.94 & \textbf{86.36} & 12.28 & 28.21 & 24.78 & 10.81 & 42.23 \\ \cmidrule(l){2-13} 
 & \multirow{3}{*}{Q2P} & GRURec & 26.02 & \textbf{23.73} & 62.46 & 83.95 & 76.25 & 21.77 & 32.04 & 17.00 & 21.62 & 40.54 \\
 &  & SRGNN & 18.76 & 22.29 & 52.94 & 84.67 & 58.72 & 21.93 & 30.34 & 13.04 & \textbf{20.86} & 35.95 \\
 &  & Attn-Mixer & 34.87 & 24.36 & 55.00 & 87.09 & 58.46 & \textbf{22.81} & 31.26 & 25.76 & 21.60 & 40.13 \\ \cmidrule(l){2-13} 
 & NQE & - & 15.82 & 11.24 & 76.79 & 87.16 & 79.52 & 11.07 & 30.76 & 11.12 & 10.14 & 37.07 \\
 & SQE-Transformer & - & 30.60 & 14.93 & 83.72 & \textbf{90.87} & 80.58 & 15.18 & 32.72 & 25.61 & 13.98 & 43.13 \\
 & SQE-LSTM & - & 31.50 & 14.10 & 83.67 & 86.70 & 84.76 & 14.46 & 30.08 & 21.92 & 12.53 & 42.19 \\ \cmidrule(l){2-13} 
 & LSGT (Ours) &  & 32.00 & 15.27 & \textbf{83.34*} & 90.61 & 86.05 & 15.62 & \textbf{33.80*} & \textbf{26.34*} & 14.45 & \textbf{44.16* (+1.03)} \\ \midrule
\multirow{10}{*}{Dressipi} & \multirow{3}{*}{FuzzQE} & GRURec & 27.62 & 94.28 & 56.15 & 77.21 & 75.40 & 94.81 & 98.43 & 23.46 & 95.52 & 71.43 \\
 &  & SRGNN & 30.18 & 94.90 & 52.41 & 74.63 & 73.38 & 95.37 & 98.32 & 25.09 & 95.69 & 71.11 \\
 &  & Attn-Mixer & 30.60 & 94.80 & 57.17 & 78.14 & 75.94 & 94.83 & \textbf{98.57} & 24.39 & 95.69 & 72.24 \\ \cmidrule(l){2-13} 
 & \multirow{3}{*}{Q2P} & GRURec & \textbf{35.93} & 95.20 & 45.22 & 66.62 & 51.20 & 96.27 & 92.58 & 25.46 & 95.45 & 67.10 \\
 &  & SRGNN & 35.48 & 95.95 & 46.05 & 64.01 & 52.58 & 95.75 & 92.81 & 25.28 & 95.68 & 67.07 \\
 &  & Attn-Mixer & 37.92 & 96.04 & 47.06 & 66.47 & 50.91 & 96.22 & 94.88 & 26.16 & 95.75 & 67.93 \\ \cmidrule(l){2-13} 
 & NQE & - & 11.52 & 95.62 & 21.19 & 52.79 & 48.28 & 96.08 & 98.04 & 13.39 & 95.80 & 59.19 \\
 & SQE-Transformer & - & 27.01 & 95.37 & 62.38 & \textbf{80.55} & \textbf{79.72} & 96.02 & 97.99 & 24.55 & 95.95 & 73.28 \\
 & SQE-LSTM & - & 25.84 & 94.81 & 62.23 & 64.19 & 70.43 & 95.39 & 96.91 & 25.23 & 95.62 & 70.07 \\ \cmidrule(l){2-13} 
 & LSGT (Ours) &  & 31.12 & \textbf{96.16*} & \textbf{64.26*} & 76.85 & 78.66 & \textbf{98.02*} & 96.98 & \textbf{28.83*} & \textbf{96.04*} & \textbf{74.10* (+0.82)} \\ \bottomrule
\end{tabular}
\end{table*}

\section{Experiment}

We use three public datasets from KDD-Cup~\footnote{https://www.aicrowd.com/challenges/amazon-kdd-cup-23-multilingual-recommendation-challenge}~\citep{jin2023amazon}, Diginetica~\footnote{https://competitions.codalab.org/competitions/11161}, and Dressipi~\footnote{https://dressipi.com/downloads/recsys-datasets} for evaluation. The number of items, sessions, and relations are reported in Table~\ref{tab:kg_details}.
Following previous work \cite{ren2020beta, wang2021benchmarking, bai2023sequential}, we use eighty percent of the edges for training, ten percent of edges for validation, and the rest of the edges as testing edges. 
As shown in Figure \ref{fig:query_types}, we conduct sampling of fourteen types of logical session queries by using the sampling algorithm described by \citep{bai2023sequential}. The number of queries is shown in Table~\ref{tab:num_queries}.
Each of the queries has a concrete meaning. For example, the 1p queries are vanilla session-based product recommendations, and the 2p queries aim to recommend product attributes based on a single session history. A detailed explanation of the query types is shown in Appendix~\ref{sess:query_type}.

\begin{table*}[t]
\centering
\small
\caption{The performance in the mean reciprocal ranking of LSGT compared with the baseline models of SQE and NQE with different backbone modules on the queries involving negations. The statistical significance is denoted by *.}
\label{tab:negation_result}
\begin{tabular}{@{}l|l|l|ccccc|l@{}}
\toprule
Dataset & Query Encoder & Session Encoder & 2ina & 2ins & 3in & inp & pin & Average Negative \\ \midrule
\multirow{10}{*}{Amazon} & \multirow{3}{*}{FuzzQE} & GRURec & 10.11 & 10.39 & 50.83 & 30.11 & 3.72 & 21.03 \\
 &  & SRGNN & 12.02 & 11.08 & 51.37 & 30.79 & 6.06 & 22.26 \\
 &  & Attn-Mixer & 17.28 & 17.47 & 53.77 & 31.96 & 4.55 & 25.00 \\ \cmidrule(l){2-9} 
 & \multirow{3}{*}{Q2P} & GRURec & 9.56 & 10.21 & 18.59 & 30.83 & 3.87 & 14.61 \\
 &  & SRGNN & 11.57 & 11.97 & 20.08 & 35.07 & 4.42 & 16.62 \\
 &  & Attn-Mixer & 18.75 & 20.68 & 51.52 & \textbf{37.04} & 6.78 & 26.95 \\ \cmidrule(l){2-9} 
 & NQE & - & 5.00 & 5.10 & 48.16 & 30.26 & 2.10 & 18.12 \\
 & SQE-Transformer & - & 18.15 & 18.88 & 55.83 & 34.76 & 8.21 & 27.16 \\
 & SQE-LSTM & - & 18.42 & 19.10 & 56.99 & 33.67 & 7.45 & 27.13 \\ \cmidrule(l){2-9} 
 & LSGT (Ours) &  & \textbf{20.98*} & \textbf{22.00*} & \textbf{60.70*} & 35.95 & \textbf{8.84*} & \textbf{29.69* (+2.93)} \\ \midrule
\multirow{10}{*}{Diginetica} & \multirow{3}{*}{FuzzQE} & GRURec & 16.15 & 9.09 & 81.65 & 14.07 & 10.69 & 26.33 \\
 &  & SRGNN & 16.62 & 15.77 & 82.30 & 14.92 & 10.69 & 28.06 \\
 &  & Attn-Mixer & 22.49 & 23.99 & 82.33 & 13.87 & 9.17 & 30.37 \\ \cmidrule(l){2-9} 
 & \multirow{3}{*}{Q2P} & GRURec & 11.42 & 9.92 & 34.33 & 10.94 & 15.58 & 16.44 \\
 &  & SRGNN & 9.17 & 8.90 & 26.28 & 11.01 & 14.84 & 14.04 \\
 &  & Attn-Mixer & 19.44 & 23.84 & 26.72 & 11.05 & 15.12 & 19.23 \\ \cmidrule(l){2-9} 
 & NQE & - & 9.71 & 11.05 & 73.10 & 11.76 & 8.60 & 22.84 \\
 & SQE-Transformer & - & 23.81 & 25.07 & 77.64 & 18.97 & 14.57 & 32.01 \\
 & SQE-LSTM & - & 23.05 & 18.56 & 81.22 & 16.77 & 13.68 & 30.66 \\ \cmidrule(l){2-9} 
 & LSGT (Ours) &  & \textbf{24.15*} & \textbf{28.69*} & \textbf{83.04*} & \textbf{19.21*} & \textbf{15.62*} & \textbf{34.14* (+2.13)} \\ \midrule
\multirow{10}{*}{Dressipi} & \multirow{3}{*}{FuzzQE} & GRURec & 20.73 & 20.97 & 50.50 & 97.37 & 92.69 & 56.45 \\
 &  & SRGNN & 23.50 & 23.68 & 50.47 & 97.36 & 92.89 & 57.58 \\
 &  & Attn-Mixer & 22.70 & 21.75 & 51.81 & 97.20 & 93.69 & 57.43 \\ \cmidrule(l){2-9} 
 & \multirow{3}{*}{Q2P} & GRURec & 20.75 & 25.64 & 24.75 & 97.97 & 63.86 & 46.59 \\
 &  & SRGNN & 20.04 & 24.35 & 26.11 & 97.70 & 64.04 & 46.45 \\
 &  & Attn-Mixer & \textbf{26.74} & \textbf{37.09} & 49.58 & \textbf{97.98} & 95.22 & 61.32 \\ \cmidrule(l){2-9} 
 & NQE & - & 8.58 & 10.60 & 14.49 & 97.40 & 94.56 & 45.13 \\
 & SQE-Transformer & - & 21.15 & 25.08 & 63.23 & 97.59 & 95.41 & 60.49 \\
 & SQE-LSTM & - & 21.03 & 24.76 & 63.14 & 97.73 & 94.50 & 60.23 \\ \cmidrule(l){2-9} 
 & LSGT (Ours) &  & 25.58 & 30.66 & \textbf{65.93*} & 97.74 & \textbf{96.30*} & \textbf{63.24* (+1.92)} \\ \bottomrule
\end{tabular}
\end{table*}

\subsection{Baseline Models}
We briefly introduce the baseline query encoding models that use various neural networks to encode the query into embedding structures. 
Here are the baseline models for the complex query-answering models: 
\begin{itemize}
    \item NQE \cite{DBLP:conf/aaai/LuoEYZGYTLW23} is a method that can be used to encode N-ary facts from the KG; 
    \item SQE \citep{bai2023sequential} uses sequence encoders to encode linearized complex queries.
\end{itemize}

In the hyper-relational session-product-attribute graph, each session can be formulated as a directed hyper-relation among various entities. Because of this, we construct the relation of \texttt{NEXT} connecting the items that are browsed in the session following the corresponding order. 
We employed a state-of-the-art session encoder to model the item history within a session. The session encoder takes into account the temporal dependencies and context of products, effectively creating a contextual representation of the entire session: 
\begin{itemize}
    \item Q2P \citep{bai-etal-2022-query2particles} uses multiple vectors to encode the queries; 
    \item FuzzQE \citep{chen2022fuzzy} use fuzzy logic to represent logical operators.
\end{itemize}

Meanwhile, the previous query encoder cannot be directly used for encoding session history as hyper-relations, so we incorporate them with session encoders.
For the session encoders, we leverage the following session encoders: 
\begin{itemize}
    \item Sequence-based encoder GRURec \citep{DBLP:conf/recsys/TanXL16};
    \item GNN-based session encoder SR-GNN \citep{DBLP:conf/aaai/WuT0WXT19};
    \item Attention-based session encoder Attention-Mixer \citep{DBLP:conf/wsdm/ZhangGLXKZXWK23}.
\end{itemize}

\subsection{Evaluation}

To precisely describe the metrics, we use the $q$ to represent a testing query and $\mathcal{G}_{val}$, $\mathcal{G}_{test}$ to represent the validation and the testing knowledge graph. Here we use $[q]_{val}$ and $[q]_{test}$ to represent the answers of query $q$ on the validation graph $\mathcal{G}_{val}$ and testing graph $\mathcal{G}_{test}$ respectively. 
Equation \ref{equa:metrics_generalization}  describes how to compute the \texttt{Inference} metrics. 
When the evaluation metric is mean reciprocal ranking (MRR), then the $m(r)$ is defined as $m(r) = \frac{1}{r}$. 
\begin{align}
   \texttt{Inference}(q) = \frac{\sum_{v \in [q]_{test}/[q]_{val}} m(\texttt{rank}(v))}{|[q]_{test}/[q]_{val}|}. \label{equa:metrics_generalization}  
\end{align}

\subsection{Experiment Details}

We maintain a consistent hidden size of $384$ for all models. This hidden size also corresponds to the size of session representation from session encoders in the baselines, as well as the query embedding size for the entire logical session query. We use the AdamW to train the models with a batch size of 512. The models are optimized with a learning rate of $0.001$, except for those with transformer structures, namely NQE, SQE-Transformer, and LSGT. These models are trained with a learning rate of $0.0001$ with a warm-up of 10000 steps. The SQE and LSGT models employ two layers of encoders. All models can be trained on GPU with 24GB memory.

\begin{table*}[h]
\centering
\caption{The ablation study on the logical structure and item orders in each session. }
\vspace{-0.3cm}
\begin{tabular}{@{}l|l|p{0.4cm}|p{0.4cm}p{0.4cm}p{0.4cm}p{0.4cm}p{0.5cm}|p{0.4cm}p{0.4cm}p{0.4cm}p{0.5cm}|p{0.4cm}p{0.4cm}p{0.4cm}p{0.4cm}p{0.4cm}@{}}
\toprule
Dataset & Encoder & \multicolumn{1}{l|}{Average} & 1p & 2p & 2ia & 2is & 3i & pi & ip & 2u & up & \multicolumn{1}{l}{2ina} & \multicolumn{1}{l}{2ins} & \multicolumn{1}{l}{3in} & \multicolumn{1}{l}{inp} & \multicolumn{1}{l}{pin} \\ \midrule
\multirow{3}{*}{Amazon} & LSGT & 31.99 & 17.73 & 9.10 & 56.73 & 84.26 & 69.39 & 19.39 & 19.47 & 15.40 & 7.86 & 20.98 & 22.00 & 60.70 & 35.95 & 8.84 \\
 & w/o Logic Structure & 15.98 & 5.41 & 2.31 & 30.31 & 50.21 & 45.21 & 3.75 & 5.49 & 4.88 & 2.56 & 16.32 & 15.77 & 38.19 & 2.13 & 1.11 \\
 & w/o Session Order & 8.45 & 6.29 & 2.59 & 17.22 & 13.85 & 19.34 & 14.07 & 3.23 & 3.49 & 1.73 & 5.50 & 4.75 & 17.54 & 4.92 & 3.73 \\ \midrule
\multirow{3}{*}{Diginetica} & LSGT & 40.59 & 32.00 & 15.27 & 83.34 & 90.61 & 86.05 & 15.62 & 33.80 & 26.34 & 14.45 & 24.15 & 28.69 & 83.04 & 19.21 & 15.62 \\
 & w/o Logic Structure & 27.17 & 18.61 & 3.84 & 68.40 & 62.80 & 64.87 & 10.13 & 20.22 & 16.08 & 8.49 & 17.38 & 14.17 & 60.37 & 9.21 & 5.74 \\
 & w/o Session Order & 17.07 & 5.08 & 9.71 & 45.49 & 34.42 & 43.23 & 9.69 & 21.71 & 3.66 & 7.92 & 4.39 & 2.56 & 35.80 & 9.98 & 5.36 \\ \midrule
\multirow{3}{*}{Dressipi} & LSGT & 70.22 & 31.12 & 96.16 & 64.26 & 76.85 & 78.66 & 98.02 & 96.98 & 28.83 & 96.04 & 25.58 & 30.66 & 65.93 & 97.74 & 96.30 \\
 & w/o Logic Structure & 25.13 & 14.87 & 2.45 & 42.03 & 59.63 & 67.62 & 9.27 & 17.71 & 18.05 & 7.64 & 19.62 & 24.67 & 59.01 & 1.95 & 7.29 \\
 & w/o Session Order & 39.78 & 9.21 & 42.80 & 21.57 & 19.57 & 23.28 & 88.31 & 61.47 & 6.31 & 68.27 & 7.41 & 6.87 & 15.96 & 96.53 & 89.35 \\ \bottomrule
\end{tabular}
\vspace{-0.3cm}

\label{Tab:ablation}
\end{table*}

\begin{table}[t]
\centering
\caption{The out-of-distribution query types evaluation. We further evaluate four types of queries with types that are unseen during the training process.}

\begin{tabular}{@{}p{1.2cm}|p{2.8cm}|p{0.4cm}p{0.4cm}p{0.4cm}p{0.6cm}|p{0.7cm}@{}}
\toprule
Dataset & Query Encoder & 3iA & 3ip & 3inA & 3inp & Ave. \\ \midrule
\multirow{5}{*}{Amazon} & FuzzQE + Attn-Mixer & 66.72 & 29.67 & 54.33 & 48.76 & 49.87 \\
 & Q2P + Attn-Mixer & 33.51 & 11.42 & 51.47 & 41.46 & 34.47 \\
 & NQE & 61.72 & 1.98 & 46.47 & 34,04 & 36.72 \\
 & SQE + Transformers & 66.03 & 28.41 & 55.61 & 51.28 & 50.33 \\
 \cmidrule{2-7}
 & LSGT (Ours) & \textbf{68.44} & \textbf{34.22} & \textbf{58.50} & \textbf{51.49} & \textbf{53.16} \\ \midrule
\multirow{5}{*}{Diginetica} & FuzzQE + Attn-Mixer & 88.30 & 32.88 & 82.75 & 34.50 & 59.61 \\
 & Q2P + Attn-Mixer & 40.28 & \textbf{43.93} & 54.31 & \textbf{48.20} & 46.68 \\
 & NQE & 86.25 & 20.79 & 64.74 & 20.93 & 48.18 \\
 & SQE + Transformers & 88.05 & 31.33 & 81.77 & 35.83 & 59.25 \\
  \cmidrule{2-7}
 & LSGT (Ours) & \textbf{91.71} & 35.24 & \textbf{83.30} & 41.05 & \textbf{62.83} \\ \midrule
\multirow{5}{*}{Dressipi} & FuzzQE + Attn-Mixer & 65.43 & 95.64 & 53.36 & 97.75 & 78.05 \\
 & Q2P + Attn-Mixer & 60.64 & 96.78 & 52.22 & 97.28 & 76.73 \\
 & NQE & 31.96 & 96.18 & 9.89 & 97.80 & 58.96 \\
 & SQE + Transformers & 72.61 & 97.12 & 55.20 & 98.14 & 80.77 \\
  \cmidrule{2-7}
 & LSGT (Ours) & \textbf{74.34} & \textbf{97.30} & \textbf{58.30} & \textbf{98.23} & \textbf{82.04} \\ \bottomrule
\end{tabular}
\vspace{-0.3cm}
\label{tab:ood}
\end{table}

\subsection{Experiment Results}

Table \ref{tab:main_result} compares the performance of different models with various backbones and configurations. 
Based on the experimental results, we can draw the following conclusions.

We found that the proposed LSGT method not only outperforms all other models but also represents the current state-of-the-art for the task. In comparison to models that solely rely on session encoders followed by query encoders, LSGT possesses the ability to leverage item information across different sessions, which proves to be critical for achieving superior performance. Additionally, LSGT demonstrates better capability in encoding graph structural inductive bias due to its operation-wise permutation invariance property, resulting in improved performance compared to other transformer-based models like SQE.

We compare the baseline of SQE \cite{bai2023sequential} in Tables \ref{tab:main_result} and \ref{tab:negation_result}, which uses a simple prefix linearization strategy \cite{DBLP:conf/iclr/LampleC20} to represent logical queries. However, because SQE is doing sequence modeling instead of graph modeling, it is not permutation invariant to logical operations, and thus cannot perform well in the query types that are sensitive to logical graph structure, like negation queries (Table  \ref{tab:negation_result}) and out-of-distribution queries (Table \ref{tab:ood}).

Furthermore, LSGT exhibits greater effectiveness in handling queries involving negations when compared to the baseline models. It achieves more significant improvements on negation queries than on EPFO queries, surpassing the performance of the best baseline.

For \texttt{2is} and \texttt{ip} query types, the any-to-any attention mechanism may not be necessary as the session embedding adequately reflects the "concentrated" intentions. However, for multi-hop questions involving negations and disjunctions, the any-to-any mechanisms demonstrate advantages.  In the Dressipi dataset, query types asking for attribute values, such as \texttt{ip}, \texttt{pi}, and others, consistently yield high performances. 
Compared to other datasets, which cover broad domains, the Dressipi dataset focuses specifically on dressing, consequently leading to a low diversity of attribute values. All the reasoning model performances reflect this property, thereby strengthening the validity of the query sampling and dataset construction.

Moreover, our observations indicate that neural models can generate more accurate results when presented with additional information or constraints. This highlights the significance of effectively modeling complex user intentions and underscores the potential for enhancing service quality in real-world usage scenarios.

\subsection{Compositional Generalization}

We conducted additional experiments on compositional generalization, and the results are presented in Table \ref{tab:ood}. In this particular setting, we evaluated the performance of our model on query types that were not encountered during the training process. These query types, namely \texttt{3iA}, \texttt{3ip}, \texttt{3inA}, and \texttt{3inp} (as illustrated in Fig.~\ref{fig:query_types}), were selected due to their complexity, involving three anchors, encompassing both EPFO queries and queries with negations, and incorporating 1-hop and 2-hop relational projections in the reasoning process. These query types were not included in the training data and were evaluated in a zero-shot manner.

By comparing the performance of our proposed method with the baselines, we observed that our approach demonstrated stronger compositional generalization on these previously unseen query types. Across the three datasets, our method improves in Mean Reciprocal Rank (MRR) ranging from 1.28 to 3.22.

\subsection{Ablation Study}

The results of the ablation study are presented in Table \ref{Tab:ablation}. In the first ablation study, we removed the tokens that represent the logical structures, and in the second ablation study, we eliminated the order information within the hypergraph by excluding the positional encoding features of item tokens in each session.
When we removed the logical structure information, a significant drop in the model's performance was observed, particularly for queries involving negations and multi-hop reasoning, such as \texttt{ip}, \texttt{pi}, \texttt{inp}, and \texttt{pin}. Without the logical structure, the model was restricted to utilizing co-occurrence information, such as "bag of sessions" and "bag of items," for ranking candidate answers. While this information may be useful for simple structured queries, its effectiveness diminished for complex structured queries.
Likewise, when we removed the order information within each session, a notable decrease in the overall performance was observed. This highlights two important findings: First, the item orders within each session play a crucial role in this task. Second, the LSGT model effectively utilizes the order information for this specific task.

\section{Conclusion}

In this paper, we presented a framework that models user intent as a complex logical query over a hyper-relational graph that describes sessions, products, and their attributes. Our framework formulates the session understanding problem as a logical session complex query answering (LS-CQA) on this graph and trains complex query-answering models to make recommendations based on the logical queries.
We also introduced a novel method of logical session graph transformer (LSGT) and demonstrated its expressiveness and operator-wise permutation invariance. Our evaluation of fourteen intersection logical reasoning tasks showed that our proposed framework achieves better results on unseen queries and queries involving negations. 
Overall, our framework provides a flexible and effective approach for modeling user intent and making recommendations in e-commerce scenarios. Future work could extend our approach to other domains and incorporate additional sources of information to improve recommendation accuracy. 

\section*{Acknowledgements}
We thank the anonymous reviewers and the area chair for their constructive comments.
The authors of this paper were supported by the NSFC Fund (U20B2053) from the NSFC of China, the RIF (R6020-19 and R6021-20), and the GRF (16211520 and 16205322) from RGC of Hong Kong. 
We also thank the support from the UGC Research Matching Grants (RMGS20EG01-D, RMGS20CR11, RMGS20CR12, RMGS20EG19, RMGS20EG21, RMGS23CR05, RMGS23EG08).

\bibliographystyle{ACM-Reference-Format}
\balance
\bibliography{sample-base}


\begin{thebibliography}{45}


\ifx \showCODEN    \undefined \def \showCODEN     #1{\unskip}     \fi
\ifx \showDOI      \undefined \def \showDOI       #1{#1}\fi
\ifx \showISBNx    \undefined \def \showISBNx     #1{\unskip}     \fi
\ifx \showISBNxiii \undefined \def \showISBNxiii  #1{\unskip}     \fi
\ifx \showISSN     \undefined \def \showISSN      #1{\unskip}     \fi
\ifx \showLCCN     \undefined \def \showLCCN      #1{\unskip}     \fi
\ifx \shownote     \undefined \def \shownote      #1{#1}          \fi
\ifx \showarticletitle \undefined \def \showarticletitle #1{#1}   \fi
\ifx \showURL      \undefined \def \showURL       {\relax}        \fi
\providecommand\bibfield[2]{#2}
\providecommand\bibinfo[2]{#2}
\providecommand\natexlab[1]{#1}
\providecommand\showeprint[2][]{arXiv:#2}

\bibitem[Alivanistos et~al\mbox{.}(2022)]%
        {DBLP:conf/iclr/AlivanistosBC022}
\bibfield{author}{\bibinfo{person}{Dimitrios Alivanistos}, \bibinfo{person}{Max Berrendorf}, \bibinfo{person}{Michael Cochez}, {and} \bibinfo{person}{Mikhail Galkin}.} \bibinfo{year}{2022}\natexlab{}.
\newblock \showarticletitle{Query Embedding on Hyper-Relational Knowledge Graphs}. In \bibinfo{booktitle}{\emph{The Tenth International Conference on Learning Representations, {ICLR} 2022, Virtual Event, April 25-29, 2022}}. \bibinfo{publisher}{OpenReview.net}.
\newblock
\urldef\tempurl%
\url{https://openreview.net/forum?id=4rLw09TgRw9}
\showURL{%
\tempurl}


\bibitem[Arakelyan et~al\mbox{.}(2021)]%
        {arakelyan2020complex}
\bibfield{author}{\bibinfo{person}{Erik Arakelyan}, \bibinfo{person}{Daniel Daza}, \bibinfo{person}{Pasquale Minervini}, {and} \bibinfo{person}{Michael Cochez}.} \bibinfo{year}{2021}\natexlab{}.
\newblock \showarticletitle{Complex Query Answering with Neural Link Predictors}. In \bibinfo{booktitle}{\emph{9th International Conference on Learning Representations, {ICLR} 2021, Virtual Event, Austria, May 3-7, 2021}}. \bibinfo{publisher}{OpenReview.net}.
\newblock
\urldef\tempurl%
\url{https://openreview.net/forum?id=Mos9F9kDwkz}
\showURL{%
\tempurl}


\bibitem[Bai et~al\mbox{.}(2023a)]%
        {DBLP:conf/nips/BaiLW0S23}
\bibfield{author}{\bibinfo{person}{Jiaxin Bai}, \bibinfo{person}{Xin Liu}, \bibinfo{person}{Weiqi Wang}, \bibinfo{person}{Chen Luo}, {and} \bibinfo{person}{Yangqiu Song}.} \bibinfo{year}{2023}\natexlab{a}.
\newblock \showarticletitle{Complex Query Answering on Eventuality Knowledge Graph with Implicit Logical Constraints}. In \bibinfo{booktitle}{\emph{Advances in Neural Information Processing Systems 36: Annual Conference on Neural Information Processing Systems 2023, NeurIPS 2023, New Orleans, LA, USA, December 10 - 16, 2023}}, \bibfield{editor}{\bibinfo{person}{Alice Oh}, \bibinfo{person}{Tristan Naumann}, \bibinfo{person}{Amir Globerson}, \bibinfo{person}{Kate Saenko}, \bibinfo{person}{Moritz Hardt}, {and} \bibinfo{person}{Sergey Levine}} (Eds.).
\newblock
\urldef\tempurl%
\url{http://papers.nips.cc/paper\_files/paper/2023/hash/6174c67b136621f3f2e4a6b1d3286f6b-Abstract-Conference.html}
\showURL{%
\tempurl}


\bibitem[Bai et~al\mbox{.}(2023b)]%
        {DBLP:journals/corr/abs-2312-13866}
\bibfield{author}{\bibinfo{person}{Jiaxin Bai}, \bibinfo{person}{Chen Luo}, \bibinfo{person}{Zheng Li}, \bibinfo{person}{Qingyu Yin}, {and} \bibinfo{person}{Yangqiu Song}.} \bibinfo{year}{2023}\natexlab{b}.
\newblock \showarticletitle{Understanding Inter-Session Intentions via Complex Logical Reasoning}.
\newblock \bibinfo{journal}{\emph{CoRR}}  \bibinfo{volume}{abs/2312.13866} (\bibinfo{year}{2023}).
\newblock
\urldef\tempurl%
\url{https://doi.org/10.48550/ARXIV.2312.13866}
\showDOI{\tempurl}
\showeprint[arXiv]{2312.13866}


\bibitem[Bai et~al\mbox{.}(2023c)]%
        {DBLP:conf/kdd/BaiLLYYS23}
\bibfield{author}{\bibinfo{person}{Jiaxin Bai}, \bibinfo{person}{Chen Luo}, \bibinfo{person}{Zheng Li}, \bibinfo{person}{Qingyu Yin}, \bibinfo{person}{Bing Yin}, {and} \bibinfo{person}{Yangqiu Song}.} \bibinfo{year}{2023}\natexlab{c}.
\newblock \showarticletitle{Knowledge Graph Reasoning over Entities and Numerical Values}. In \bibinfo{booktitle}{\emph{Proceedings of the 29th {ACM} {SIGKDD} Conference on Knowledge Discovery and Data Mining, {KDD} 2023, Long Beach, CA, USA, August 6-10, 2023}}, \bibfield{editor}{\bibinfo{person}{Ambuj~K. Singh}, \bibinfo{person}{Yizhou Sun}, \bibinfo{person}{Leman Akoglu}, \bibinfo{person}{Dimitrios Gunopulos}, \bibinfo{person}{Xifeng Yan}, \bibinfo{person}{Ravi Kumar}, \bibinfo{person}{Fatma Ozcan}, {and} \bibinfo{person}{Jieping Ye}} (Eds.). \bibinfo{publisher}{{ACM}}, \bibinfo{pages}{57--68}.
\newblock
\urldef\tempurl%
\url{https://doi.org/10.1145/3580305.3599399}
\showDOI{\tempurl}


\bibitem[Bai et~al\mbox{.}(2022b)]%
        {bai-etal-2022-query2particles}
\bibfield{author}{\bibinfo{person}{Jiaxin Bai}, \bibinfo{person}{Zihao Wang}, \bibinfo{person}{Hongming Zhang}, {and} \bibinfo{person}{Yangqiu Song}.} \bibinfo{year}{2022}\natexlab{b}.
\newblock \showarticletitle{{Q}uery2{P}articles: Knowledge Graph Reasoning with Particle Embeddings}. In \bibinfo{booktitle}{\emph{Findings of the Association for Computational Linguistics: NAACL 2022}}. \bibinfo{publisher}{Association for Computational Linguistics}, \bibinfo{address}{Seattle, United States}, \bibinfo{pages}{2703--2714}.
\newblock
\urldef\tempurl%
\url{https://doi.org/10.18653/v1/2022.findings-naacl.207}
\showDOI{\tempurl}


\bibitem[Bai et~al\mbox{.}(2023d)]%
        {bai2023sequential}
\bibfield{author}{\bibinfo{person}{Jiaxin Bai}, \bibinfo{person}{Tianshi Zheng}, {and} \bibinfo{person}{Yangqiu Song}.} \bibinfo{year}{2023}\natexlab{d}.
\newblock \showarticletitle{Sequential Query Encoding for Complex Query Answering on Knowledge Graphs}.
\newblock \bibinfo{journal}{\emph{Transactions on Machine Learning Research}} (\bibinfo{year}{2023}).
\newblock
\showISSN{2835-8856}
\urldef\tempurl%
\url{https://openreview.net/forum?id=ERqGqZzSu5}
\showURL{%
\tempurl}


\bibitem[Bai et~al\mbox{.}(2022a)]%
        {bai2022answering}
\bibfield{author}{\bibinfo{person}{Yushi Bai}, \bibinfo{person}{Xin Lv}, \bibinfo{person}{Juanzi Li}, {and} \bibinfo{person}{Lei Hou}.} \bibinfo{year}{2022}\natexlab{a}.
\newblock \showarticletitle{Answering Complex Logical Queries on Knowledge Graphs via Query Computation Tree Optimization}.
\newblock \bibinfo{journal}{\emph{CoRR}}  \bibinfo{volume}{abs/2212.09567} (\bibinfo{year}{2022}).
\newblock
\urldef\tempurl%
\url{https://doi.org/10.48550/arXiv.2212.09567}
\showDOI{\tempurl}
\showeprint[arXiv]{2212.09567}


\bibitem[Barcel{\'{o}} et~al\mbox{.}(2022)]%
        {DBLP:conf/log/Barcelo00O22}
\bibfield{author}{\bibinfo{person}{Pablo Barcel{\'{o}}}, \bibinfo{person}{Mikhail Galkin}, \bibinfo{person}{Christopher Morris}, {and} \bibinfo{person}{Miguel A.~Romero Orth}.} \bibinfo{year}{2022}\natexlab{}.
\newblock \showarticletitle{Weisfeiler and Leman Go Relational}. In \bibinfo{booktitle}{\emph{Learning on Graphs Conference, LoG 2022, 9-12 December 2022, Virtual Event}} \emph{(\bibinfo{series}{Proceedings of Machine Learning Research}, Vol.~\bibinfo{volume}{198})}, \bibfield{editor}{\bibinfo{person}{Bastian Rieck} {and} \bibinfo{person}{Razvan Pascanu}} (Eds.). \bibinfo{publisher}{{PMLR}}, \bibinfo{pages}{46}.
\newblock
\urldef\tempurl%
\url{https://proceedings.mlr.press/v198/barcelo22a.html}
\showURL{%
\tempurl}


\bibitem[Chen et~al\mbox{.}(2022)]%
        {chen2022fuzzy}
\bibfield{author}{\bibinfo{person}{Xuelu Chen}, \bibinfo{person}{Ziniu Hu}, {and} \bibinfo{person}{Yizhou Sun}.} \bibinfo{year}{2022}\natexlab{}.
\newblock \showarticletitle{Fuzzy Logic Based Logical Query Answering on Knowledge Graphs}. In \bibinfo{booktitle}{\emph{Thirty-Sixth {AAAI} Conference on Artificial Intelligence, {AAAI} 2022, Thirty-Fourth Conference on Innovative Applications of Artificial Intelligence, {IAAI} 2022, The Twelveth Symposium on Educational Advances in Artificial Intelligence, {EAAI} 2022 Virtual Event, February 22 - March 1, 2022}}. \bibinfo{publisher}{{AAAI} Press}, \bibinfo{pages}{3939--3948}.
\newblock
\urldef\tempurl%
\url{https://ojs.aaai.org/index.php/AAAI/article/view/20310}
\showURL{%
\tempurl}


\bibitem[Guo et~al\mbox{.}(2022)]%
        {DBLP:conf/cikm/GuoZL0ZK22}
\bibfield{author}{\bibinfo{person}{Jiayan Guo}, \bibinfo{person}{Peiyan Zhang}, \bibinfo{person}{Chaozhuo Li}, \bibinfo{person}{Xing Xie}, \bibinfo{person}{Yan Zhang}, {and} \bibinfo{person}{Sunghun Kim}.} \bibinfo{year}{2022}\natexlab{}.
\newblock \showarticletitle{Evolutionary Preference Learning via Graph Nested {GRU} {ODE} for Session-based Recommendation}. In \bibinfo{booktitle}{\emph{Proceedings of the 31st {ACM} International Conference on Information {\&} Knowledge Management, Atlanta, GA, USA, October 17-21, 2022}}, \bibfield{editor}{\bibinfo{person}{Mohammad~Al Hasan} {and} \bibinfo{person}{Li~Xiong}} (Eds.). \bibinfo{publisher}{{ACM}}, \bibinfo{pages}{624--634}.
\newblock
\urldef\tempurl%
\url{https://doi.org/10.1145/3511808.3557314}
\showDOI{\tempurl}


\bibitem[Hamilton et~al\mbox{.}(2018)]%
        {hamilton2018embedding}
\bibfield{author}{\bibinfo{person}{William~L. Hamilton}, \bibinfo{person}{Payal Bajaj}, \bibinfo{person}{Marinka Zitnik}, \bibinfo{person}{Dan Jurafsky}, {and} \bibinfo{person}{Jure Leskovec}.} \bibinfo{year}{2018}\natexlab{}.
\newblock \showarticletitle{Embedding Logical Queries on Knowledge Graphs}. In \bibinfo{booktitle}{\emph{Advances in Neural Information Processing Systems 31: Annual Conference on Neural Information Processing Systems 2018, NeurIPS 2018, December 3-8, 2018, Montr{\'{e}}al, Canada}}, \bibfield{editor}{\bibinfo{person}{Samy Bengio}, \bibinfo{person}{Hanna~M. Wallach}, \bibinfo{person}{Hugo Larochelle}, \bibinfo{person}{Kristen Grauman}, \bibinfo{person}{Nicol{\`{o}} Cesa{-}Bianchi}, {and} \bibinfo{person}{Roman Garnett}} (Eds.). \bibinfo{pages}{2030--2041}.
\newblock
\urldef\tempurl%
\url{https://proceedings.neurips.cc/paper/2018/hash/ef50c335cca9f340bde656363ebd02fd-Abstract.html}
\showURL{%
\tempurl}


\bibitem[Hidasi et~al\mbox{.}(2015)]%
        {hidasi2015session}
\bibfield{author}{\bibinfo{person}{Bal{\'a}zs Hidasi}, \bibinfo{person}{Alexandros Karatzoglou}, \bibinfo{person}{Linas Baltrunas}, {and} \bibinfo{person}{Domonkos Tikk}.} \bibinfo{year}{2015}\natexlab{}.
\newblock \showarticletitle{Session-based recommendations with recurrent neural networks}.
\newblock \bibinfo{journal}{\emph{arXiv preprint arXiv:1511.06939}} (\bibinfo{year}{2015}).
\newblock


\bibitem[Hu et~al\mbox{.}(2024)]%
        {DBLP:journals/corr/abs-2402-14609}
\bibfield{author}{\bibinfo{person}{Qi Hu}, \bibinfo{person}{Weifeng Jiang}, \bibinfo{person}{Haoran Li}, \bibinfo{person}{Zihao Wang}, \bibinfo{person}{Jiaxin Bai}, \bibinfo{person}{Qianren Mao}, \bibinfo{person}{Yangqiu Song}, \bibinfo{person}{Lixin Fan}, {and} \bibinfo{person}{Jianxin Li}.} \bibinfo{year}{2024}\natexlab{}.
\newblock \showarticletitle{FedCQA: Answering Complex Queries on Multi-Source Knowledge Graphs via Federated Learning}.
\newblock \bibinfo{journal}{\emph{CoRR}}  \bibinfo{volume}{abs/2402.14609} (\bibinfo{year}{2024}).
\newblock
\urldef\tempurl%
\url{https://doi.org/10.48550/ARXIV.2402.14609}
\showDOI{\tempurl}
\showeprint[arXiv]{2402.14609}


\bibitem[Hu et~al\mbox{.}(2023)]%
        {DBLP:journals/corr/abs-2312-15591}
\bibfield{author}{\bibinfo{person}{Qi Hu}, \bibinfo{person}{Haoran Li}, \bibinfo{person}{Jiaxin Bai}, {and} \bibinfo{person}{Yangqiu Song}.} \bibinfo{year}{2023}\natexlab{}.
\newblock \showarticletitle{Privacy-Preserving Neural Graph Databases}.
\newblock \bibinfo{journal}{\emph{CoRR}}  \bibinfo{volume}{abs/2312.15591} (\bibinfo{year}{2023}).
\newblock
\urldef\tempurl%
\url{https://doi.org/10.48550/ARXIV.2312.15591}
\showDOI{\tempurl}
\showeprint[arXiv]{2312.15591}


\bibitem[Huang et~al\mbox{.}(2023)]%
        {DBLP:journals/corr/abs-2302-02209}
\bibfield{author}{\bibinfo{person}{Xingyue Huang}, \bibinfo{person}{Miguel A.~Romero Orth}, \bibinfo{person}{{\.I}smail~{\.I}lkan Ceylan}, {and} \bibinfo{person}{Pablo Barcel{\'{o}}}.} \bibinfo{year}{2023}\natexlab{}.
\newblock \showarticletitle{A Theory of Link Prediction via Relational Weisfeiler-Leman}.
\newblock \bibinfo{journal}{\emph{CoRR}}  \bibinfo{volume}{abs/2302.02209} (\bibinfo{year}{2023}).
\newblock
\urldef\tempurl%
\url{https://doi.org/10.48550/ARXIV.2302.02209}
\showDOI{\tempurl}
\showeprint[arXiv]{2302.02209}


\bibitem[Huang et~al\mbox{.}(2022)]%
        {DBLP:conf/icml/HuangWLH22}
\bibfield{author}{\bibinfo{person}{Zhongyu Huang}, \bibinfo{person}{Yingheng Wang}, \bibinfo{person}{Chaozhuo Li}, {and} \bibinfo{person}{Huiguang He}.} \bibinfo{year}{2022}\natexlab{}.
\newblock \showarticletitle{Going Deeper into Permutation-Sensitive Graph Neural Networks}. In \bibinfo{booktitle}{\emph{International Conference on Machine Learning, {ICML} 2022, 17-23 July 2022, Baltimore, Maryland, {USA}}} \emph{(\bibinfo{series}{Proceedings of Machine Learning Research}, Vol.~\bibinfo{volume}{162})}, \bibfield{editor}{\bibinfo{person}{Kamalika Chaudhuri}, \bibinfo{person}{Stefanie Jegelka}, \bibinfo{person}{Le~Song}, \bibinfo{person}{Csaba Szepesv{\'{a}}ri}, \bibinfo{person}{Gang Niu}, {and} \bibinfo{person}{Sivan Sabato}} (Eds.). \bibinfo{publisher}{{PMLR}}, \bibinfo{pages}{9377--9409}.
\newblock
\urldef\tempurl%
\url{https://proceedings.mlr.press/v162/huang22l.html}
\showURL{%
\tempurl}


\bibitem[Jin et~al\mbox{.}(2023)]%
        {jin2023amazon}
\bibfield{author}{\bibinfo{person}{Wei Jin}, \bibinfo{person}{Haitao Mao}, \bibinfo{person}{Zheng Li}, \bibinfo{person}{Haoming Jiang}, \bibinfo{person}{Chen Luo}, \bibinfo{person}{Hongzhi Wen}, \bibinfo{person}{Haoyu Han}, \bibinfo{person}{Hanqing Lu}, \bibinfo{person}{Zhengyang Wang}, \bibinfo{person}{Ruirui Li}, {et~al\mbox{.}}} \bibinfo{year}{2023}\natexlab{}.
\newblock \showarticletitle{Amazon-M2: A Multilingual Multi-locale Shopping Session Dataset for Recommendation and Text Generation}.
\newblock \bibinfo{journal}{\emph{arXiv preprint arXiv:2307.09688}} (\bibinfo{year}{2023}).
\newblock


\bibitem[Kim et~al\mbox{.}(2022)]%
        {DBLP:conf/nips/KimNMCLLH22}
\bibfield{author}{\bibinfo{person}{Jinwoo Kim}, \bibinfo{person}{Dat Nguyen}, \bibinfo{person}{Seonwoo Min}, \bibinfo{person}{Sungjun Cho}, \bibinfo{person}{Moontae Lee}, \bibinfo{person}{Honglak Lee}, {and} \bibinfo{person}{Seunghoon Hong}.} \bibinfo{year}{2022}\natexlab{}.
\newblock \showarticletitle{Pure Transformers are Powerful Graph Learners}. In \bibinfo{booktitle}{\emph{NeurIPS}}.
\newblock
\urldef\tempurl%
\url{http://papers.nips.cc/paper\_files/paper/2022/hash/5d84236751fe6d25dc06db055a3180b0-Abstract-Conference.html}
\showURL{%
\tempurl}


\bibitem[Lample and Charton(2020)]%
        {DBLP:conf/iclr/LampleC20}
\bibfield{author}{\bibinfo{person}{Guillaume Lample} {and} \bibinfo{person}{Fran{\c{c}}ois Charton}.} \bibinfo{year}{2020}\natexlab{}.
\newblock \showarticletitle{Deep Learning For Symbolic Mathematics}. In \bibinfo{booktitle}{\emph{8th International Conference on Learning Representations, {ICLR} 2020, Addis Ababa, Ethiopia, April 26-30, 2020}}. \bibinfo{publisher}{OpenReview.net}.
\newblock
\urldef\tempurl%
\url{https://openreview.net/forum?id=S1eZYeHFDS}
\showURL{%
\tempurl}


\bibitem[Li et~al\mbox{.}(2021)]%
        {DBLP:conf/sigir/LiPLSLXYCZZ21}
\bibfield{author}{\bibinfo{person}{Chaozhuo Li}, \bibinfo{person}{Bochen Pang}, \bibinfo{person}{Yuming Liu}, \bibinfo{person}{Hao Sun}, \bibinfo{person}{Zheng Liu}, \bibinfo{person}{Xing Xie}, \bibinfo{person}{Tianqi Yang}, \bibinfo{person}{Yanling Cui}, \bibinfo{person}{Liangjie Zhang}, {and} \bibinfo{person}{Qi Zhang}.} \bibinfo{year}{2021}\natexlab{}.
\newblock \showarticletitle{AdsGNN: Behavior-Graph Augmented Relevance Modeling in Sponsored Search}. In \bibinfo{booktitle}{\emph{{SIGIR} '21: The 44th International {ACM} {SIGIR} Conference on Research and Development in Information Retrieval, Virtual Event, Canada, July 11-15, 2021}}, \bibfield{editor}{\bibinfo{person}{Fernando Diaz}, \bibinfo{person}{Chirag Shah}, \bibinfo{person}{Torsten Suel}, \bibinfo{person}{Pablo Castells}, \bibinfo{person}{Rosie Jones}, {and} \bibinfo{person}{Tetsuya Sakai}} (Eds.). \bibinfo{publisher}{{ACM}}, \bibinfo{pages}{223--232}.
\newblock
\urldef\tempurl%
\url{https://doi.org/10.1145/3404835.3462926}
\showDOI{\tempurl}


\bibitem[Li et~al\mbox{.}(2017)]%
        {li2017neural}
\bibfield{author}{\bibinfo{person}{Jing Li}, \bibinfo{person}{Pengjie Ren}, \bibinfo{person}{Zhumin Chen}, \bibinfo{person}{Zhaochun Ren}, \bibinfo{person}{Tao Lian}, {and} \bibinfo{person}{Jun Ma}.} \bibinfo{year}{2017}\natexlab{}.
\newblock \showarticletitle{Neural attentive session-based recommendation}. In \bibinfo{booktitle}{\emph{Proceedings of the 2017 ACM on Conference on Information and Knowledge Management}}. \bibinfo{pages}{1419--1428}.
\newblock


\bibitem[Liu et~al\mbox{.}(2024)]%
        {DBLP:conf/www/LiuWBST24}
\bibfield{author}{\bibinfo{person}{Lihui Liu}, \bibinfo{person}{Zihao Wang}, \bibinfo{person}{Jiaxin Bai}, \bibinfo{person}{Yangqiu Song}, {and} \bibinfo{person}{Hanghang Tong}.} \bibinfo{year}{2024}\natexlab{}.
\newblock \showarticletitle{New Frontiers of Knowledge Graph Reasoning: Recent Advances and Future Trends}. In \bibinfo{booktitle}{\emph{Companion Proceedings of the {ACM} on Web Conference 2024, {WWW} 2024, Singapore, Singapore, May 13-17, 2024}}, \bibfield{editor}{\bibinfo{person}{Tat{-}Seng Chua}, \bibinfo{person}{Chong{-}Wah Ngo}, \bibinfo{person}{Roy~Ka{-}Wei Lee}, \bibinfo{person}{Ravi Kumar}, {and} \bibinfo{person}{Hady~W. Lauw}} (Eds.). \bibinfo{publisher}{{ACM}}, \bibinfo{pages}{1294--1297}.
\newblock
\urldef\tempurl%
\url{https://doi.org/10.1145/3589335.3641254}
\showDOI{\tempurl}


\bibitem[Liu et~al\mbox{.}(2018)]%
        {DBLP:conf/kdd/LiuZMZ18}
\bibfield{author}{\bibinfo{person}{Qiao Liu}, \bibinfo{person}{Yifu Zeng}, \bibinfo{person}{Refuoe Mokhosi}, {and} \bibinfo{person}{Haibin Zhang}.} \bibinfo{year}{2018}\natexlab{}.
\newblock \showarticletitle{{STAMP:} Short-Term Attention/Memory Priority Model for Session-based Recommendation}. In \bibinfo{booktitle}{\emph{Proceedings of the 24th {ACM} {SIGKDD} International Conference on Knowledge Discovery {\&} Data Mining, {KDD} 2018, London, UK, August 19-23, 2018}}, \bibfield{editor}{\bibinfo{person}{Yike Guo} {and} \bibinfo{person}{Faisal Farooq}} (Eds.). \bibinfo{publisher}{{ACM}}, \bibinfo{pages}{1831--1839}.
\newblock
\urldef\tempurl%
\url{https://doi.org/10.1145/3219819.3219950}
\showDOI{\tempurl}


\bibitem[Liu et~al\mbox{.}(2022)]%
        {liu2022mask}
\bibfield{author}{\bibinfo{person}{Xiao Liu}, \bibinfo{person}{Shiyu Zhao}, \bibinfo{person}{Kai Su}, \bibinfo{person}{Yukuo Cen}, \bibinfo{person}{Jiezhong Qiu}, \bibinfo{person}{Mengdi Zhang}, \bibinfo{person}{Wei Wu}, \bibinfo{person}{Yuxiao Dong}, {and} \bibinfo{person}{Jie Tang}.} \bibinfo{year}{2022}\natexlab{}.
\newblock \showarticletitle{Mask and Reason: Pre-Training Knowledge Graph Transformers for Complex Logical Queries}. In \bibinfo{booktitle}{\emph{{KDD} '22: The 28th {ACM} {SIGKDD} Conference on Knowledge Discovery and Data Mining, Washington, DC, USA, August 14 - 18, 2022}}, \bibfield{editor}{\bibinfo{person}{Aidong Zhang} {and} \bibinfo{person}{Huzefa Rangwala}} (Eds.). \bibinfo{publisher}{{ACM}}, \bibinfo{pages}{1120--1130}.
\newblock
\urldef\tempurl%
\url{https://doi.org/10.1145/3534678.3539472}
\showDOI{\tempurl}


\bibitem[Luo et~al\mbox{.}(2023)]%
        {DBLP:conf/aaai/LuoEYZGYTLW23}
\bibfield{author}{\bibinfo{person}{Haoran Luo}, \bibinfo{person}{Haihong E}, \bibinfo{person}{Yuhao Yang}, \bibinfo{person}{Gengxian Zhou}, \bibinfo{person}{Yikai Guo}, \bibinfo{person}{Tianyu Yao}, \bibinfo{person}{Zichen Tang}, \bibinfo{person}{Xueyuan Lin}, {and} \bibinfo{person}{Kaiyang Wan}.} \bibinfo{year}{2023}\natexlab{}.
\newblock \showarticletitle{{NQE:} N-ary Query Embedding for Complex Query Answering over Hyper-Relational Knowledge Graphs}. In \bibinfo{booktitle}{\emph{Thirty-Seventh {AAAI} Conference on Artificial Intelligence, {AAAI} 2023, Thirty-Fifth Conference on Innovative Applications of Artificial Intelligence, {IAAI} 2023, Thirteenth Symposium on Educational Advances in Artificial Intelligence, {EAAI} 2023, Washington, DC, USA, February 7-14, 2023}}, \bibfield{editor}{\bibinfo{person}{Brian Williams}, \bibinfo{person}{Yiling Chen}, {and} \bibinfo{person}{Jennifer Neville}} (Eds.). \bibinfo{publisher}{{AAAI} Press}, \bibinfo{pages}{4543--4551}.
\newblock
\urldef\tempurl%
\url{https://ojs.aaai.org/index.php/AAAI/article/view/25576}
\showURL{%
\tempurl}


\bibitem[Maron et~al\mbox{.}(2019)]%
        {DBLP:conf/iclr/MaronBSL19}
\bibfield{author}{\bibinfo{person}{Haggai Maron}, \bibinfo{person}{Heli Ben{-}Hamu}, \bibinfo{person}{Nadav Shamir}, {and} \bibinfo{person}{Yaron Lipman}.} \bibinfo{year}{2019}\natexlab{}.
\newblock \showarticletitle{Invariant and Equivariant Graph Networks}. In \bibinfo{booktitle}{\emph{7th International Conference on Learning Representations, {ICLR} 2019, New Orleans, LA, USA, May 6-9, 2019}}. \bibinfo{publisher}{OpenReview.net}.
\newblock
\urldef\tempurl%
\url{https://openreview.net/forum?id=Syx72jC9tm}
\showURL{%
\tempurl}


\bibitem[Pan et~al\mbox{.}(2020)]%
        {DBLP:conf/cikm/PanCCCR20}
\bibfield{author}{\bibinfo{person}{Zhiqiang Pan}, \bibinfo{person}{Fei Cai}, \bibinfo{person}{Wanyu Chen}, \bibinfo{person}{Honghui Chen}, {and} \bibinfo{person}{Maarten de Rijke}.} \bibinfo{year}{2020}\natexlab{}.
\newblock \showarticletitle{Star Graph Neural Networks for Session-based Recommendation}. In \bibinfo{booktitle}{\emph{{CIKM} '20: The 29th {ACM} International Conference on Information and Knowledge Management, Virtual Event, Ireland, October 19-23, 2020}}, \bibfield{editor}{\bibinfo{person}{Mathieu d'Aquin}, \bibinfo{person}{Stefan Dietze}, \bibinfo{person}{Claudia Hauff}, \bibinfo{person}{Edward Curry}, {and} \bibinfo{person}{Philippe Cudr{\'{e}}{-}Mauroux}} (Eds.). \bibinfo{publisher}{{ACM}}, \bibinfo{pages}{1195--1204}.
\newblock
\urldef\tempurl%
\url{https://doi.org/10.1145/3340531.3412014}
\showDOI{\tempurl}


\bibitem[Ren et~al\mbox{.}(2020)]%
        {ren2020query2box}
\bibfield{author}{\bibinfo{person}{Hongyu Ren}, \bibinfo{person}{Weihua Hu}, {and} \bibinfo{person}{Jure Leskovec}.} \bibinfo{year}{2020}\natexlab{}.
\newblock \showarticletitle{Query2box: Reasoning over Knowledge Graphs in Vector Space Using Box Embeddings}. In \bibinfo{booktitle}{\emph{8th International Conference on Learning Representations, {ICLR} 2020, Addis Ababa, Ethiopia, April 26-30, 2020}}. \bibinfo{publisher}{OpenReview.net}.
\newblock
\urldef\tempurl%
\url{https://openreview.net/forum?id=BJgr4kSFDS}
\showURL{%
\tempurl}


\bibitem[Ren and Leskovec(2020)]%
        {ren2020beta}
\bibfield{author}{\bibinfo{person}{Hongyu Ren} {and} \bibinfo{person}{Jure Leskovec}.} \bibinfo{year}{2020}\natexlab{}.
\newblock \showarticletitle{Beta Embeddings for Multi-Hop Logical Reasoning in Knowledge Graphs}. In \bibinfo{booktitle}{\emph{Advances in Neural Information Processing Systems 33: Annual Conference on Neural Information Processing Systems 2020, NeurIPS 2020, December 6-12, 2020, virtual}}, \bibfield{editor}{\bibinfo{person}{Hugo Larochelle}, \bibinfo{person}{Marc'Aurelio Ranzato}, \bibinfo{person}{Raia Hadsell}, \bibinfo{person}{Maria{-}Florina Balcan}, {and} \bibinfo{person}{Hsuan{-}Tien Lin}} (Eds.).
\newblock
\urldef\tempurl%
\url{https://proceedings.neurips.cc/paper/2020/hash/e43739bba7cdb577e9e3e4e42447f5a5-Abstract.html}
\showURL{%
\tempurl}


\bibitem[Tan et~al\mbox{.}(2016)]%
        {DBLP:conf/recsys/TanXL16}
\bibfield{author}{\bibinfo{person}{Yong~Kiam Tan}, \bibinfo{person}{Xinxing Xu}, {and} \bibinfo{person}{Yong Liu}.} \bibinfo{year}{2016}\natexlab{}.
\newblock \showarticletitle{Improved Recurrent Neural Networks for Session-based Recommendations}. In \bibinfo{booktitle}{\emph{Proceedings of the 1st Workshop on Deep Learning for Recommender Systems, DLRS@RecSys 2016, Boston, MA, USA, September 15, 2016}}, \bibfield{editor}{\bibinfo{person}{Alexandros Karatzoglou}, \bibinfo{person}{Bal{\'{a}}zs Hidasi}, \bibinfo{person}{Domonkos Tikk}, \bibinfo{person}{Oren~Sar Shalom}, \bibinfo{person}{Haggai Roitman}, \bibinfo{person}{Bracha Shapira}, {and} \bibinfo{person}{Lior Rokach}} (Eds.). \bibinfo{publisher}{{ACM}}, \bibinfo{pages}{17--22}.
\newblock
\urldef\tempurl%
\url{https://doi.org/10.1145/2988450.2988452}
\showDOI{\tempurl}


\bibitem[Tang and Wang(2018)]%
        {DBLP:conf/wsdm/TangW18}
\bibfield{author}{\bibinfo{person}{Jiaxi Tang} {and} \bibinfo{person}{Ke Wang}.} \bibinfo{year}{2018}\natexlab{}.
\newblock \showarticletitle{Personalized Top-N Sequential Recommendation via Convolutional Sequence Embedding}. In \bibinfo{booktitle}{\emph{Proceedings of the Eleventh {ACM} International Conference on Web Search and Data Mining, {WSDM} 2018, Marina Del Rey, CA, USA, February 5-9, 2018}}, \bibfield{editor}{\bibinfo{person}{Yi~Chang}, \bibinfo{person}{Chengxiang Zhai}, \bibinfo{person}{Yan Liu}, {and} \bibinfo{person}{Yoelle Maarek}} (Eds.). \bibinfo{publisher}{{ACM}}, \bibinfo{pages}{565--573}.
\newblock
\urldef\tempurl%
\url{https://doi.org/10.1145/3159652.3159656}
\showDOI{\tempurl}


\bibitem[Tang et~al\mbox{.}(2023)]%
        {DBLP:conf/sigir/TangFTP0S23}
\bibfield{author}{\bibinfo{person}{Zhenwei Tang}, \bibinfo{person}{Griffin Floto}, \bibinfo{person}{Armin Toroghi}, \bibinfo{person}{Shichao Pei}, \bibinfo{person}{Xiangliang Zhang}, {and} \bibinfo{person}{Scott Sanner}.} \bibinfo{year}{2023}\natexlab{}.
\newblock \showarticletitle{LogicRec: Recommendation with Users' Logical Requirements}. In \bibinfo{booktitle}{\emph{Proceedings of the 46th International {ACM} {SIGIR} Conference on Research and Development in Information Retrieval, {SIGIR} 2023, Taipei, Taiwan, July 23-27, 2023}}, \bibfield{editor}{\bibinfo{person}{Hsin{-}Hsi Chen}, \bibinfo{person}{Wei{-}Jou~(Edward) Duh}, \bibinfo{person}{Hen{-}Hsen Huang}, \bibinfo{person}{Makoto~P. Kato}, \bibinfo{person}{Josiane Mothe}, {and} \bibinfo{person}{Barbara Poblete}} (Eds.). \bibinfo{publisher}{{ACM}}, \bibinfo{pages}{2129--2133}.
\newblock
\urldef\tempurl%
\url{https://doi.org/10.1145/3539618.3592012}
\showDOI{\tempurl}


\bibitem[Vashishth et~al\mbox{.}(2020)]%
        {Vashishth2020Composition-based}
\bibfield{author}{\bibinfo{person}{Shikhar Vashishth}, \bibinfo{person}{Soumya Sanyal}, \bibinfo{person}{Vikram Nitin}, {and} \bibinfo{person}{Partha Talukdar}.} \bibinfo{year}{2020}\natexlab{}.
\newblock \showarticletitle{Composition-based Multi-Relational Graph Convolutional Networks}. In \bibinfo{booktitle}{\emph{International Conference on Learning Representations}}.
\newblock
\urldef\tempurl%
\url{https://openreview.net/forum?id=BylA_C4tPr}
\showURL{%
\tempurl}


\bibitem[Wang et~al\mbox{.}(2023a)]%
        {DBLP:conf/acl/WangFYSWS23}
\bibfield{author}{\bibinfo{person}{Zihao Wang}, \bibinfo{person}{Weizhi Fei}, \bibinfo{person}{Hang Yin}, \bibinfo{person}{Yangqiu Song}, \bibinfo{person}{Ginny~Y. Wong}, {and} \bibinfo{person}{Simon See}.} \bibinfo{year}{2023}\natexlab{a}.
\newblock \showarticletitle{Wasserstein-Fisher-Rao Embedding: Logical Query Embeddings with Local Comparison and Global Transport}. In \bibinfo{booktitle}{\emph{Findings of the Association for Computational Linguistics: {ACL} 2023, Toronto, Canada, July 9-14, 2023}}, \bibfield{editor}{\bibinfo{person}{Anna Rogers}, \bibinfo{person}{Jordan~L. Boyd{-}Graber}, {and} \bibinfo{person}{Naoaki Okazaki}} (Eds.). \bibinfo{publisher}{Association for Computational Linguistics}, \bibinfo{pages}{13679--13696}.
\newblock
\urldef\tempurl%
\url{https://doi.org/10.18653/V1/2023.FINDINGS-ACL.864}
\showDOI{\tempurl}


\bibitem[Wang et~al\mbox{.}(2023b)]%
        {DBLP:journals/corr/abs-2301-08859}
\bibfield{author}{\bibinfo{person}{Zihao Wang}, \bibinfo{person}{Yangqiu Song}, \bibinfo{person}{Ginny~Y. Wong}, {and} \bibinfo{person}{Simon See}.} \bibinfo{year}{2023}\natexlab{b}.
\newblock \showarticletitle{Logical Message Passing Networks with One-hop Inference on Atomic Formulas}. In \bibinfo{booktitle}{\emph{The Eleventh International Conference on Learning Representations, {ICLR} 2023, Kigali, Rwanda, May 1-5, 2023}}. \bibinfo{publisher}{OpenReview.net}.
\newblock
\urldef\tempurl%
\url{https://openreview.net/pdf?id=SoyOsp7i\_l}
\showURL{%
\tempurl}


\bibitem[Wang et~al\mbox{.}(2021)]%
        {wang2021benchmarking}
\bibfield{author}{\bibinfo{person}{Zihao Wang}, \bibinfo{person}{Hang Yin}, {and} \bibinfo{person}{Yangqiu Song}.} \bibinfo{year}{2021}\natexlab{}.
\newblock \showarticletitle{Benchmarking the Combinatorial Generalizability of Complex Query Answering on Knowledge Graphs}. In \bibinfo{booktitle}{\emph{Proceedings of the Neural Information Processing Systems Track on Datasets and Benchmarks 1, NeurIPS Datasets and Benchmarks 2021, December 2021, virtual}}, \bibfield{editor}{\bibinfo{person}{Joaquin Vanschoren} {and} \bibinfo{person}{Sai{-}Kit Yeung}} (Eds.).
\newblock
\urldef\tempurl%
\url{https://datasets-benchmarks-proceedings.neurips.cc/paper/2021/hash/7eabe3a1649ffa2b3ff8c02ebfd5659f-Abstract-round2.html}
\showURL{%
\tempurl}


\bibitem[Wu et~al\mbox{.}(2019)]%
        {DBLP:conf/aaai/WuT0WXT19}
\bibfield{author}{\bibinfo{person}{Shu Wu}, \bibinfo{person}{Yuyuan Tang}, \bibinfo{person}{Yanqiao Zhu}, \bibinfo{person}{Liang Wang}, \bibinfo{person}{Xing Xie}, {and} \bibinfo{person}{Tieniu Tan}.} \bibinfo{year}{2019}\natexlab{}.
\newblock \showarticletitle{Session-Based Recommendation with Graph Neural Networks}. In \bibinfo{booktitle}{\emph{The Thirty-Third {AAAI} Conference on Artificial Intelligence, {AAAI} 2019, The Thirty-First Innovative Applications of Artificial Intelligence Conference, {IAAI} 2019, The Ninth {AAAI} Symposium on Educational Advances in Artificial Intelligence, {EAAI} 2019, Honolulu, Hawaii, USA, January 27 - February 1, 2019}}. \bibinfo{publisher}{{AAAI} Press}, \bibinfo{pages}{346--353}.
\newblock
\urldef\tempurl%
\url{https://doi.org/10.1609/aaai.v33i01.3301346}
\showDOI{\tempurl}


\bibitem[Xia et~al\mbox{.}(2021)]%
        {DBLP:conf/aaai/0013YYWC021}
\bibfield{author}{\bibinfo{person}{Xin Xia}, \bibinfo{person}{Hongzhi Yin}, \bibinfo{person}{Junliang Yu}, \bibinfo{person}{Qinyong Wang}, \bibinfo{person}{Lizhen Cui}, {and} \bibinfo{person}{Xiangliang Zhang}.} \bibinfo{year}{2021}\natexlab{}.
\newblock \showarticletitle{Self-Supervised Hypergraph Convolutional Networks for Session-based Recommendation}. In \bibinfo{booktitle}{\emph{Thirty-Fifth {AAAI} Conference on Artificial Intelligence, {AAAI} 2021, Thirty-Third Conference on Innovative Applications of Artificial Intelligence, {IAAI} 2021, The Eleventh Symposium on Educational Advances in Artificial Intelligence, {EAAI} 2021, Virtual Event, February 2-9, 2021}}. \bibinfo{publisher}{{AAAI} Press}, \bibinfo{pages}{4503--4511}.
\newblock
\urldef\tempurl%
\url{https://ojs.aaai.org/index.php/AAAI/article/view/16578}
\showURL{%
\tempurl}


\bibitem[Xu et~al\mbox{.}(2019)]%
        {DBLP:conf/iclr/XuHLJ19}
\bibfield{author}{\bibinfo{person}{Keyulu Xu}, \bibinfo{person}{Weihua Hu}, \bibinfo{person}{Jure Leskovec}, {and} \bibinfo{person}{Stefanie Jegelka}.} \bibinfo{year}{2019}\natexlab{}.
\newblock \showarticletitle{How Powerful are Graph Neural Networks?}. In \bibinfo{booktitle}{\emph{7th International Conference on Learning Representations, {ICLR} 2019, New Orleans, LA, USA, May 6-9, 2019}}. \bibinfo{publisher}{OpenReview.net}.
\newblock
\urldef\tempurl%
\url{https://openreview.net/forum?id=ryGs6iA5Km}
\showURL{%
\tempurl}


\bibitem[Xu et~al\mbox{.}(2022)]%
        {xu2022neural}
\bibfield{author}{\bibinfo{person}{Zezhong Xu}, \bibinfo{person}{Wen Zhang}, \bibinfo{person}{Peng Ye}, \bibinfo{person}{Hui Chen}, {and} \bibinfo{person}{Huajun Chen}.} \bibinfo{year}{2022}\natexlab{}.
\newblock \showarticletitle{Neural-Symbolic Entangled Framework for Complex Query Answering}. In \bibinfo{booktitle}{\emph{Advances in Neural Information Processing Systems}}, \bibfield{editor}{\bibinfo{person}{S.~Koyejo}, \bibinfo{person}{S.~Mohamed}, \bibinfo{person}{A.~Agarwal}, \bibinfo{person}{D.~Belgrave}, \bibinfo{person}{K.~Cho}, {and} \bibinfo{person}{A.~Oh}} (Eds.), Vol.~\bibinfo{volume}{35}. \bibinfo{publisher}{Curran Associates, Inc.}, \bibinfo{pages}{1806--1819}.
\newblock
\urldef\tempurl%
\url{https://proceedings.neurips.cc/paper_files/paper/2022/file/0bcfb525c8f8f07ae10a93d0b2a40e00-Paper-Conference.pdf}
\showURL{%
\tempurl}


\bibitem[Yang et~al\mbox{.}(2022)]%
        {yang2022gammae}
\bibfield{author}{\bibinfo{person}{Dong Yang}, \bibinfo{person}{Peijun Qing}, \bibinfo{person}{Yang Li}, \bibinfo{person}{Haonan Lu}, {and} \bibinfo{person}{Xiaodong Lin}.} \bibinfo{year}{2022}\natexlab{}.
\newblock \showarticletitle{GammaE: Gamma Embeddings for Logical Queries on Knowledge Graphs}. In \bibinfo{booktitle}{\emph{Proceedings of the 2022 Conference on Empirical Methods in Natural Language Processing, {EMNLP} 2022, Abu Dhabi, United Arab Emirates, December 7-11, 2022}}, \bibfield{editor}{\bibinfo{person}{Yoav Goldberg}, \bibinfo{person}{Zornitsa Kozareva}, {and} \bibinfo{person}{Yue Zhang}} (Eds.). \bibinfo{publisher}{Association for Computational Linguistics}, \bibinfo{pages}{745--760}.
\newblock
\urldef\tempurl%
\url{https://aclanthology.org/2022.emnlp-main.47}
\showURL{%
\tempurl}


\bibitem[Yin et~al\mbox{.}(2023)]%
        {DBLP:journals/corr/abs-2307-13701}
\bibfield{author}{\bibinfo{person}{Hang Yin}, \bibinfo{person}{Zihao Wang}, \bibinfo{person}{Weizhi Fei}, {and} \bibinfo{person}{Yangqiu Song}.} \bibinfo{year}{2023}\natexlab{}.
\newblock \showarticletitle{EFO\({}_{\mbox{k}}\)-CQA: Towards Knowledge Graph Complex Query Answering beyond Set Operation}.
\newblock \bibinfo{journal}{\emph{CoRR}}  \bibinfo{volume}{abs/2307.13701} (\bibinfo{year}{2023}).
\newblock
\urldef\tempurl%
\url{https://doi.org/10.48550/ARXIV.2307.13701}
\showDOI{\tempurl}
\showeprint[arXiv]{2307.13701}


\bibitem[Zhang et~al\mbox{.}(2023)]%
        {DBLP:conf/wsdm/ZhangGLXKZXWK23}
\bibfield{author}{\bibinfo{person}{Peiyan Zhang}, \bibinfo{person}{Jiayan Guo}, \bibinfo{person}{Chaozhuo Li}, \bibinfo{person}{Yueqi Xie}, \bibinfo{person}{Jaeboum Kim}, \bibinfo{person}{Yan Zhang}, \bibinfo{person}{Xing Xie}, \bibinfo{person}{Haohan Wang}, {and} \bibinfo{person}{Sunghun Kim}.} \bibinfo{year}{2023}\natexlab{}.
\newblock \showarticletitle{Efficiently Leveraging Multi-level User Intent for Session-based Recommendation via Atten-Mixer Network}. In \bibinfo{booktitle}{\emph{Proceedings of the Sixteenth {ACM} International Conference on Web Search and Data Mining, {WSDM} 2023, Singapore, 27 February 2023 - 3 March 2023}}, \bibfield{editor}{\bibinfo{person}{Tat{-}Seng Chua}, \bibinfo{person}{Hady~W. Lauw}, \bibinfo{person}{Luo Si}, \bibinfo{person}{Evimaria Terzi}, {and} \bibinfo{person}{Panayiotis Tsaparas}} (Eds.). \bibinfo{publisher}{{ACM}}, \bibinfo{pages}{168--176}.
\newblock
\urldef\tempurl%
\url{https://doi.org/10.1145/3539597.3570445}
\showDOI{\tempurl}


\bibitem[Zhu et~al\mbox{.}(2022)]%
        {zhu2022neural}
\bibfield{author}{\bibinfo{person}{Zhaocheng Zhu}, \bibinfo{person}{Mikhail Galkin}, \bibinfo{person}{Zuobai Zhang}, {and} \bibinfo{person}{Jian Tang}.} \bibinfo{year}{2022}\natexlab{}.
\newblock \showarticletitle{Neural-Symbolic Models for Logical Queries on Knowledge Graphs}. In \bibinfo{booktitle}{\emph{International Conference on Machine Learning, {ICML} 2022, 17-23 July 2022, Baltimore, Maryland, {USA}}} \emph{(\bibinfo{series}{Proceedings of Machine Learning Research}, Vol.~\bibinfo{volume}{162})}, \bibfield{editor}{\bibinfo{person}{Kamalika Chaudhuri}, \bibinfo{person}{Stefanie Jegelka}, \bibinfo{person}{Le~Song}, \bibinfo{person}{Csaba Szepesv{\'{a}}ri}, \bibinfo{person}{Gang Niu}, {and} \bibinfo{person}{Sivan Sabato}} (Eds.). \bibinfo{publisher}{{PMLR}}, \bibinfo{pages}{27454--27478}.
\newblock
\urldef\tempurl%
\url{https://proceedings.mlr.press/v162/zhu22c.html}
\showURL{%
\tempurl}


\end{thebibliography}

\begin{table}[]
\centering
\tiny
\caption{The query types and their explanations. }
\begin{tabular}{@{}l|p{7cm}@{}}
\toprule
Query Types &  Explanations \\ \midrule
1p & Predict the product that is desired by a given session. \\   \midrule
2p & { Predict} the attribute value of the product that is desired by a given session. \\ \midrule
2iA & Predict the product that is desired by a given session with a certain attribute value. \\ \midrule
2iS & Predict the product that is desired by both given sessions. \\ \midrule
3i & Predict the product that is desired by both given sessions with a certain attribute value. \\  \midrule
ip & Predict the attribute value of the { product} that is desired by both of {  the given sessions}. \\ \midrule
pi & Predict the attribute value of the { product} that is desired by { a given session}, this attribute value is possessed by another given item. \\ \midrule
2u & Predict the product that is desired by either one of the sessions. \\ \midrule
up & Predict the attribute value of the product that is desired by either of the sessions. \\ \midrule
2inA & Predict the product that is desired by a given session, but does not have a certain attribute. \\ \midrule
2inS & Predict the product that is desired by a given session, but is not wanted by another session. \\ \midrule
3in & Predict the product that is desired by a given session with { a certain} attribute, but is not wanted by another session. \\ \midrule
inp & Predict the attribute value of the product that is desired by a given session, but is not wanted by another session. \\ \midrule
pin & Predict the attribute value of the product that is desired by a given session, but is not possessed by another given item. \\ \bottomrule
\end{tabular}
\label{tab:type_explanation}
\end{table}

\appendix
\newpage
\section{Proofs}
\label{sess:proof}
We give the proofs of \textbf{Theorem} \ref{theorem:expressiveness}, \ref{theorem:expressiveness_RWL}, and  \textbf{Theorem} \ref{theorem:permutation_invariance}. Before proving these two theorems, we define the proxy graph of the  graph we used in this paper that involves N-ary facts. 

\paragraph{Defintion (Proxy Graph)} For each computational graph utilized by the query encoder, we can uniquely identify the corresponding proxy graph. This graph comprises binary edges without hyperedges and consists of vertices representing items, sessions, and operators. The edges in the proxy graph can be categorized into three types: session edges, which connect item vertices to session vertices and utilize their position, such as $1, 2, \dots, k$ as edge types; relational projection edges, which connect two vertices and employ the relation type as the edge type; and logical edges, which utilize the corresponding logical operation type as the edge type.
It is important to note that the proxy graph is distinct for different computational graphs with N-ary facts.

 \paragraph{Defintion (Non-relational Argumented Proxy  Graph)} 

For each proxy graph,  we create another graph called a Non-relational Argumented Proxy Graph. This graph includes all vertices in the original proxy. Meanwhile, the argument graph an additional node for each edge in the original graph, and it takes relation type as a node feature.  

\begin{lemma} \label{lemma:baseline_message_passing}
    Encoding the complex session query by following the computational graph using a session encoder followed by query encoding is equivalent to performing message passing on the corresponding proxy graph.
\end{lemma}
\begin{proof} 
To prove this, we must analyze each operation in the original N-ary computational graph. For the session encoder part, the session representation is computed from the items it contains, which is equivalent to a message passing on the proxy graph with a unique aggregation function, namely the session encoder.
For the intersection and union operations, the computational graph utilizes various specially designed logical operations to encode them, and they can be considered as messages passing over the proxy graph.
Similarly, for the relational projection, the tail node aggregates information from the head node and relation type, which is also a message-passing process on the proxy graph.
\end{proof}

\begin{lemma} \label{lemma:lsgt_tokengt}
    The encoding process of LSGT is equivalent to using TokenGT to encode the proxy graph. 
\end{lemma}

\begin{proof} 
The encoding process of LSGT consists of three parts. First, the node tokens are used to identify and represent the items, sessions, and operators. Secondly, the logical structure tokens are employed to represent the logical connections between items and sessions. Finally, LSGT utilizes positional embedding as the token feature to describe the positional information of an item in a session. This process is equivalent to building an edge between the item and session and assigning its edge feature as the corresponding position embedding, which is done in the proxy graph. Therefore, encoding logical session graphs using LSGT is equivalent to using TokenGT on the proxy graph.
\end{proof}

\begin{lemma} \label{lemma:graph_equivalence.}
Suppose the $G_1$ and $G_2$ are two proxy graphs, and $G_1'$ and $G_2'$ are two non-relational argument proxy graphs converted from $G_1$ and $G_2$ respectively. Then $G_1 = G_2 \longleftrightarrow G_1' = G_2'$.
\end{lemma}
\begin{proof} 
The direction  $G_1 = G_2 \rightarrow G_1' = G_2'$ is trivial because according to the definition, the conversion process is deterministic. We focus on the reverse side: $G_1 = G_2 \leftarrow G_1' = G_2'$. We try to prove it by contradiction. Suppose $G_1 \neq G_2$ but $G_1' = G_2'$. Without losing generality, we can suppose there is an edge $(u, v, r) \in G_1$ but $(u, v, r)$ is not in $G_2$ where $u, v$ are vertices and $r$ is the relation. Because of this, suppose $w$ is a node with feature $r$ connected that is linked to both $u, v$ in the argument graph for both $G_1' = G_2'$. Namely both $(u,w)$ and $(w,v)$ are in  $G_1' = G_2'$. Because the $(u, v, r)$ is not in $G_2$, $(w, v)$ is not constructed by the edge $(u,v,r)$, thus it must be constructed by another edge $(u', v, r)$. This suggests $w$ is connected with at least three vertices $u, v$ and $u'$. This is contradictory to the definition of the non-relational argument proxy graph. 
\end{proof}

\paragraph{Proof of Theorem \ref{theorem:expressiveness}}
\begin{proof}

Based on Lemma \ref{lemma:baseline_message_passing}, as the baseline models perform message passing on the proxy graph, their expressiveness is as powerful as the 1-WL graph isomorphism test \citep{DBLP:conf/iclr/XuHLJ19}.
Additionally, according to Lemma \ref{lemma:baseline_message_passing}, the encoding process of LSGT on the session query graph is equivalent to using order-2 TokenGT on the proxy graph. Order-2 TokenGT can approximate the 2-IGN network \citep{DBLP:conf/nips/KimNMCLLH22}, and the 2-IGN network is at least as expressive as the 2-WL graph isomorphism test \citep{DBLP:conf/iclr/MaronBSL19}.
Since the 2-WL test is equivalent to the 1-WL test, we can conclude that LSGT has at least the same expressiveness as the baseline models.

\end{proof}

\begin{table}[]
\centering
\small
\caption{
The query structures are used for training and evaluation. For brevity, the $p$, $i$, $n$, and $u$ represent the projection, intersection, negation, and union operations. The query types are trained and evaluated under supervised settings. 
}
\begin{tabular}{@{}lcccc@{}}
\toprule
 & \multicolumn{2}{c}{Train Queries} & Validation Queries & Test Queries \\ \midrule
Dataset & Item-Attribute & \multicolumn{1}{l}{Others} & All Types & All Types \\
Amazon & 2,535,506 & 720,816 & 36,041 & 36,041 \\
Diginetica & 249,562 & 60,235 & 3,012 & 3,012 \\
Dressipi & 414,083 & 668,650 & 33,433 & 33,433 \\ \bottomrule
\end{tabular}
\label{tab:num_queries}
\end{table}

\paragraph{Proof of Theorem \ref{theorem:expressiveness_RWL}}
\begin{proof}
To prove the expressiveness of LSGT on the multirelational proxy graph is at least 1-RWL, we need to show that for two non-isomorphic multi-relational graphs $G$ and $H$, if they can be distinguished by 1-RWL or equivalently CompGCN, then it also can be distinguished by LSGT. 

According to the CompGCN definition and the definition of the Non-Relational Argument Proxy Graph of $G'$ and $H'$ which are constructed from $G$ and $H$ respectively, CompGCN computed on $G$ and $H$ can be regarded as a message passing on the non-relational message passing on $G'$ and $H'$. We will give a more detailed justification as follows. 

The formula of CompGCN \cite{Vashishth2020Composition-based} is as follows:

\begin{equation}
    h_v^{k+1} = f(\sum_{(u,r)\in N(v)} W^k_{\lambda (r)}\phi(h_u^k, h_r^k)),
    \label{equa:compgcn}
\end{equation}

here $h_u^k$, $h_r^k$ denotes features for node $u$ and relation $r$ at the $k$-th layer respectively, $h_v^{k+1}$ denotes the updated
representation of node $v$, and $W_{\lambda(r)} \in  \mathbb{R}^{d \times d} $  
is a relation-type specific parameter. 
$f$ is an activation function (such as the \texttt{ReLU}).
In CompGCN,
we use direction-specific weights, i.e.,  $\lambda(r)  = dir(r)$.

On the other hand, the general form of non-relational message passing is expressed as follows:

\begin{equation}
    h_v^{k+1} = \gamma^{k+1}(h_v^{k}, \bigoplus_{u \in N(v)} \psi^{k+1} (h_v^{k}, h_u^{k})), 
    \label{equa:mpnn}
\end{equation}
Where denotes in the $k$-th layer, the $\bigoplus$ is a differentiable, permutation invariant function, e.g., sum, mean or max, and $\gamma$ and $\psi $ denote differentiable functions such as MLPs (Multi Layer Perceptrons).

With these two formulas we are going to prove the CompGCN computed on relational graph $G$ with Equation (\ref{equa:compgcn}) is equivalent to a message passing on the non-relational argumented graph $G'$ with Equation (\ref{equa:mpnn}). To prove this, we can use a constructive method to show that each step of CompGCN with equation (\ref{equa:compgcn}) on $G$ is identical to two steps message passing on $G'$ with equation (\ref{equa:mpnn}).
Suppose for each edge $(u, v)$ in graph $G$, there is a in-between node $w$ with node label of relation type $r$ in the augmented graph $G'$. 
The computation of equation \ref{equa:compgcn} can be separated into the following two steps.
First for each neighbor $u$ of $v$, we use $g^k_{u,r}$ to represent the result of the result of a composition of the following two functions: 
\begin{align}
    g^k_{u,r} &= \phi(h_u^k, h_r^k) \label{equa:compgcn_step_1} \\
    h_v^{k+1} &= f(\sum_{(u,r)\in N(v)} W^k_{\lambda (r)} g^k_{u,r})
    \label{equa:compgcn_step_2} 
\end{align}

Then we are going to prove that these two steps each equivalent to one message passing step on the argumented graph $G'$.
First we show the equation \ref{equa:compgcn_step_1} is equivalent to a  message passing process on $G'$ to the augmented node $w$ that contains the relation feature $h_w = h_r$ between $u$ and $v$. We can let the $\gamma^{k+1}(x, y) = y$ and $\psi^{k+1}(x, y) = \phi(y, x)$ in the general message passing in equation \ref{equa:mpnn}, and we got the following equation
\begin{align}
    h_w^{k+1} &= \gamma^{k+1}(h_w^{k}, \bigoplus_{u \in N(w)} \psi^{k+1} (h_w^{k}, h_u^{k})) \\
            &= \bigoplus_{u \in N(w)} \psi^{k+1} (h_w^{k}, h_u^{k}) \\
            &= \psi^{k+1} (h_w^{k}, h_u^{k}) = \phi(h_u^{k}, h_w^{k}) \\
            &= \phi(h_u^{k}, h_r^{k}) = g^k_{u,r}
\end{align}
    
Then, we show that in the second step of CompGCN in equation \ref{equa:compgcn_step_2} is equivalent to another step of message passing in $G'$ from each of its neighbors $w$ to $v$. 
According to the Equation \ref{equa:mpnn}, we can write
\begin{align}
    h_v^{k+1} &= \gamma^{k+1}(h_v^{k}, \bigoplus_{u \in N(w)} \psi^{k+1} (h_w^{k}, h_u^{k})) \\
    &= f(\bigoplus_{u \in N(w)} \psi^{k+1} (h_w^{k}, h_u^{k})) \\
    &= f(\sum_{u \in N(w)} \psi^{k+1} (h_w^{k}, h_u^{k})) \\
    &= f(\sum_{u \in N(w)} W^k_{\lambda (r)} h_w^{k} ) \\
    &= f(\sum_{u \in N(w)} W^k_{\lambda (r)} g^k_{u,r} ).
\end{align}
As a result, the CompGCN over $G$ is able to write as two steps of message passing over the augmented graph $G'$.

Thus, if $G$ and $H$ can be distinguished by CompGCN, then $G'$ and $H'$ can be distinguished by a certain non-relational message-passing algorithm. Thus $G'$ and $H'$ can be distinguished by the 1-WL test.
As Shown in previous proof LSGT is at least as powerful as the 2-WL test, and 1-WL and 2-WL tests are equivalent. We can conclude that LSGT is able to distinguish $G'$ and $H'$. 
According to Lemma \ref{lemma:graph_equivalence.}, if LSGT is able to distinguish $G'$ and $H'$ then it is able to distinguish $G$ and $H$.

\end{proof}

\paragraph{Proof of Theorem \ref{theorem:permutation_invariance}}
\begin{proof}
Operation-wise permutation invariance mainly focuses on the \texttt{Intersection} and \texttt{Union} operations. Suppose the input vertices for such an operator are $\{p_1, p_2, \dots, p_n\}$. If an arbitrary permutation over these vertices is denoted as $\{p_1', p_2', \dots, p_n'\}$, a global permutation of token identifiers can be constructed, where vertices $p_i$ are mapped to $p_i'$ and the rest are mapped to themselves. 
As per Lemma \ref{lemma:lsgt_tokengt}, LSGT can approximate 2-IGN \citep{DBLP:conf/iclr/MaronBSL19}, which is permutation invariant. Therefore, LSGT can approximate a query encoder that achieves operation-wise permutation invariance.
\end{proof}

\section{The Concrete Meanings of Various Query Types}
\label{sess:query_type}
In this session, we describe concrete meanings of the query types shown in Figure~\ref{fig:query_types}. The meanings are listed in the Table~\ref{tab:type_explanation}.

\end{document}